\documentclass[runningheads]{llncs}

\usepackage{eccv}

\usepackage{eccvabbrv}

\usepackage{graphicx}
\usepackage{booktabs}

\usepackage[accsupp]{axessibility}  

\usepackage[pagebackref,breaklinks,colorlinks,citecolor=eccvblue]{hyperref}

\usepackage{orcidlink}
\usepackage{marvosym}
\begin{document}

\title{ZMIS-SAM: Segment Anything Model Enhanced with Wavelet Transform for Zooplankton Microscopy Image Instance Segmentation} 

\titlerunning{ZMIS-SAM}

\author{Dekun Yuan\inst{1}\orcidlink{0000-0003-2597-6395} \and
Zhongwei Li\inst{1, \textsuperscript{\Letter}}\orcidlink{0000-0002-3934-9053} \and
Zheng Qiao\inst{1}\orcidlink{0009-0001-3151-8967} \and
Jie Zhang\inst{1}\orcidlink{0000-0002-2635-7783}}

\authorrunning{Dekun.Y et al.}

\institute{$^{1}$College of Oceanography and Space Informatics, China University of Petroleum (East China), Qingdao, 266580, China\\
\email{ydk\_libra0903@163.com, li.zhongwei@vip.163.com}\\
}

\maketitle

\begin{abstract}
As primary consumers in the marine food chain, zooplankton play a crucial role in maintaining marine ecological balance. However, the Segment Anything Model (SAM) exhibits limited performance in microscopic image instance segmentation due to its lack of zooplankton-specific domain knowledge. To address these challenges, we propose a novel instance segmentation model based on SAM and wavelet transform (ZMIS-SAM), effectively tackling issues such as inaccurate classification, discontinuous segmentation of slender appendages, and incomplete boundary segmentation. Our framework incorporates three core innovations: ZM-ViT enhances SAM's capability to model zooplankton morphology and image intensity distributions through two lightweight adapters, the Neighboring Feature Aggregation Module (NFAM) improves continuous segmentation of semi-transparent slender appendages by integrating general-purpose and domain-specific features, and the Wavelet-based Multi-scale Multi-directional Feature Enhancement (WM2FE) module effectively recovers high-frequency details to refine boundary segmentation completeness. Extensive experiments demonstrate that ZMIS-SAM achieves state-of-the-art instance segmentation performance on the zooplankton dataset and exhibits strong generalization capability across multiple public cross-domain datasets. Code: \href{https://github.com/sdydk/ZMIS-SAM}{ZMIS-SAM}.
\end{abstract}

\section{Introduction}
\label{sec:intro}

Zooplankton, serving as a crucial intermediate link in the marine food chain, plays a vital role in maintaining the stability of marine ecosystems\cite{Zooplankton}. Current methods for zooplankton identification can be broadly categorized into manual and intelligent approaches\cite{10752977, 11104797, 10595451}. Traditional manual identification relies on biological experts observing and counting specimens under microscopes. While this method achieves high accuracy, it suffers from low efficiency and falls short of meeting the demands for efficient, large-scale analysis of zooplankton communities\cite{Pu_2021_ICCV, YE2026112719}. With the advancement of artificial intelligence technologies, intelligent identification techniques based on deep learning methods, such as instance segmentation, offer a promising alternative for achieving efficient and accurate zooplankton identification.

Instance segmentation stands as one of the most challenging tasks in intelligent visual recognition, requiring not only pixel-wise dense prediction but also distinguishing between different object instances within the same category\cite{9356353}. Two-stage methods, exemplified by Mask R-CNN \cite{MaskRCNN}, have become the mainstream paradigm in this field. These approaches explicitly generate region proposals and perform dense prediction to produce instance-level masks. While they demonstrate stronger segmentation performance on domain-specific datasets compared to single-stage methods, their effectiveness heavily depends on dataset scale, image quality, and the distribution and quantity of object instances. This characteristic makes it difficult for two-stage methods to achieve satisfactory instance segmentation performance in the special field of zooplankton. With their powerful zero-shot generalization capabilities learned from vast and diverse datasets, recent advances in large-scale pre-trained models \cite{radford21a, SAM, ravi2024sam2} can extract transferable visual representations, providing a more efficient and flexible paradigm, for example, segmentation in specialized domains like zooplankton research.

\begin{figure}[!t]
\centering
\includegraphics[width=3.5in]{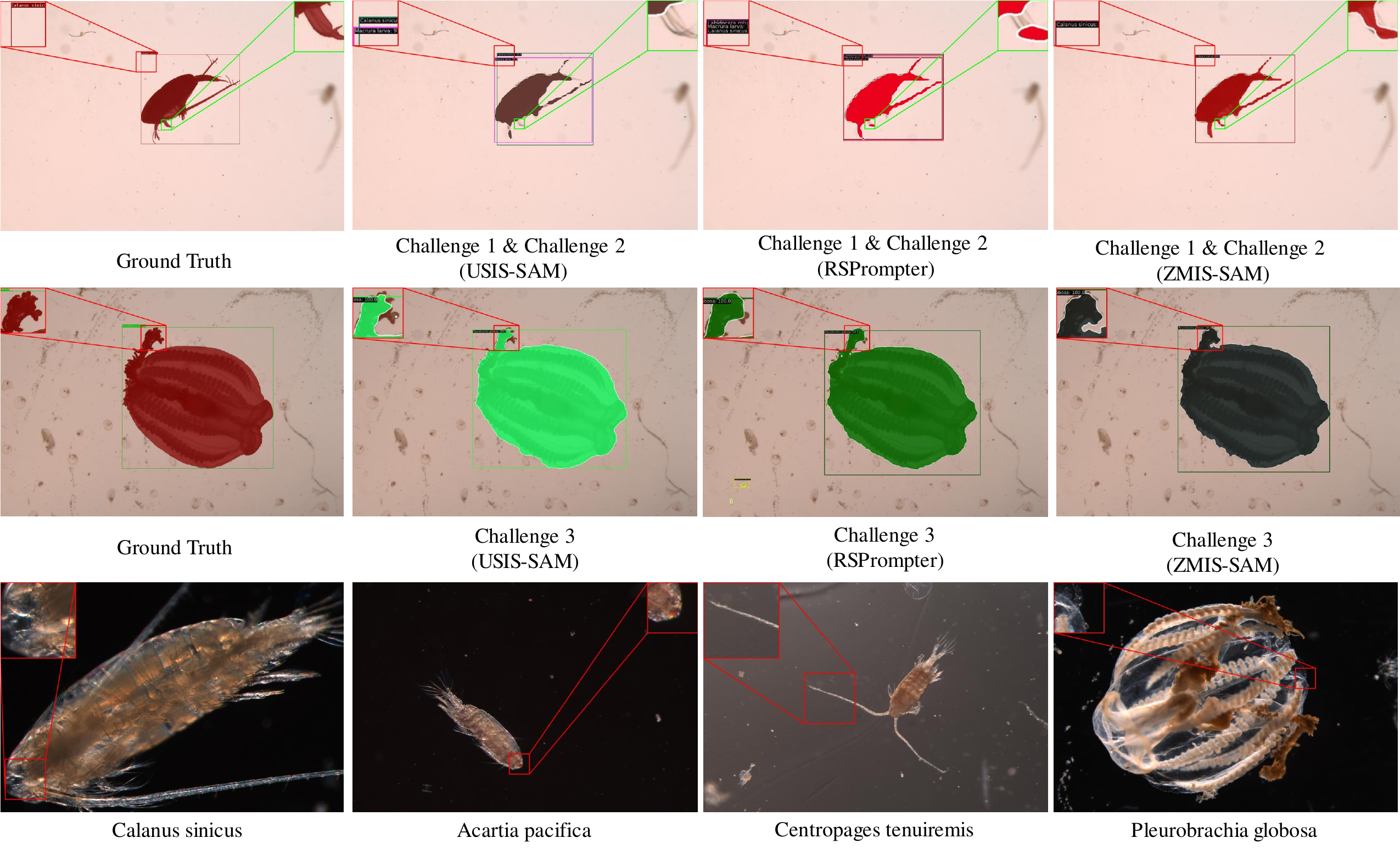}
\caption{Comparison of ZMIS-SAM with other state-of-the-art models on the ZMIS5K dataset. Challenge 1, Challenge 2, and Challenge 3 represent inaccurate classification, discontinuous segmentation of slender appendages, and incomplete boundary segmentation, respectively. }
\label{fig: Challenges}
\end{figure}

The Segment Anything Model (SAM)\cite{SAM} has emerged as a highly popular large-scale vision foundation model with strong zero-shot segmentation capabilities, allowing it to be effectively adapted to various downstream tasks such as image inpainting \cite{yu2023inpaint, 10657905} and instance segmentation\cite{USIS-SAM, RSPrompter, FoodSAM}. For example, USIS-SAM \cite{USIS-SAM} and RSPrompter \cite{RSPrompter} have successfully applied SAM to instance segmentation in underwater and remote sensing imagery, respectively, providing valuable references for its extension to vertical domains. However, due to significant differences in morphological characteristics between zooplankton and typical underwater or remote sensing targets, as well as distinct imaging mechanisms in microscopy, adapting SAM to zooplankton instance segmentation poses several challenges, as illustrated in \cref{fig: Challenges}: 
\begin{itemize}
  \item [1)] \textbf{Inaccurate Classification:} Zooplankton encompass a wide variety of species, and organisms from different genera or families often exhibit highly similar morphological appearances under the microscope. For instance, as shown in the third row of \cref{fig: Challenges}, \emph{Calanus sinicus} and \emph{Acartia pacifica} have similar body sizes across different magnification levels, with the key distinguishing feature being the red eye spot present in \emph{Acartia pacifica}. However, due to the lack of fine-grained prior knowledge about zooplankton, SAM tends to misclassify \emph{Acartia pacifica} as \emph{Calanus sinicus} \cite{Zhang2023ACS}.
  \item [2)] \textbf{Discontinuous Segmentation of Slender Appendages:} Some zooplankton possess translucent slender appendages. \emph{Centropages tenuiremis} uses two elongated, semi-transparent antennae on its head for feeding and sensing environmental changes. However, existing SAM-based instance segmentation models fail to fully leverage features extracted by the image encoder, resulting in fragmented predictions of these fine structures.
  \item [3)]\textbf{Incomplete Boundary Segmentation:} To enhance predation efficiency or avoid predators, some zooplankton have evolved transparent body structures, such as \emph{Pleurobrachia globosa}. When adapting SAM for instance segmentation in such specialized domains, a common approach involves upsampling low-resolution feature maps to recover high-resolution details and improve model robustness. However, the upsampling process often fails to fully restore high-frequency details, making it difficult for the model to accurately separate these translucent organisms from the background, ultimately leading to incomplete boundary segmentation \cite{2016Image, 8099589}.
\end{itemize}

To address the aforementioned challenges, we constructed the first high-quality zooplankton microscopy image instance segmentation dataset (ZMIS5K) using imaging from a research-grade stereomicroscope. ZMIS5K comprises 5,358 pixel-level annotated images across 47 species, containing 10,228 instances in total.  Furthermore, we propose a SAM-based instance segmentation model specifically designed for zooplankton microscopy images. To address the issues encountered when applying SAM to zooplankton instance segmentation, the innovations of our proposed ZMIS‑SAM are as follows: \textbf{1)} For classification inaccuracy, we design two novel adapter modules, Shape Adapter (SA) and Intensity Adapter (IA), built upon the widely-used lightweight adapter tuning framework \cite{9879683}. These two adapters, together with ViT \cite{dosovitskiy2020vit}, constitute ZM-ViT. These are designed to capture zooplankton-specific morphological characteristics and microscopy-specific channel intensity distributions, respectively. The proposed ZM-ViT integrates these adapters into selected ViT within SAM's image encoder, enabling effective domain-specific feature learning. \textbf{2)} For discontinuous segmentation of slender appendages, we design a Neighboring Feature Aggregation Module (NFAM) that effectively integrates general-purpose features from frozen ViT layers in SAM's image encoder and domain-specific features from ZM-ViT. This enhances the model's ability to segment translucent slender appendages with improved continuity. \textbf{3)} To tackle incomplete boundary segmentation, we design a Wavelet-based Multi-scale and Multi-directional Feature Enhancement (WM2FE) module. The resulting feature enhancement module compensates for high-frequency details lost during upsampling, thereby improving segmentation accuracy along transparent or low-contrast edges. The main contributions of this work are as follows:

\begin{itemize}
  \item [1)] To our knowledge, ZMIS-SAM is the first large vision model for instance segmentation in zooplankton microscopy images. It shows strong performance in ZMIS5K and offers valuable insight for adapting SAM to other vertical domains.
  \item [2)] We propose three key components: ZM-ViT enhances domain-specific representation learning for improved species classification, an NFAM improves segmentation continuity of translucent slender appendages by combining general and domain-specific features, and a WM2FE module recovers high-frequency details to refine boundary segmentation.
  \item [3)] We introduce ZMIS5K, the species-rich, high-resolution zooplankton microscopy image dataset for instance segmentation. It includes 5,358 images across 47 species with 10,228 instances, and supports various vision tasks such as image classification, object detection, semantic segmentation, and instance segmentation. For a more detailed introduction to the ZMIS5K dataset, please refer to \textbf{Appendix A.}
\end{itemize}

\section{Related work}
\subsection{Instance Segmentation}
Instance segmentation methods are primarily categorized into single-stage and two-stage approaches. Single-stage methods \cite{yolact-iccv2019, wang2020solo, chen2020blendmask, PointRend} simultaneously predict object category, location, and segmentation mask to achieve rapid instance segmentation, though their segmentation accuracy is lower than that of two-stage methods. Two-stage methods utilize object detection results for segmentation mask tasks, divided into bottom-up and top-down approaches. Bottom-up methods \cite{ISTR, BottomupIS, Lazarow_2022_CVPR} first perform pixel-level predictions, then employ clustering methods to distinguish different instances. This approach better preserves low-level information such as details, but its poor generalization capability makes it challenging to handle complex scenes\cite{WaterMask}. Top-down approaches \cite{MaskRCNN, Mask2former, Maskformer, R3} first generate bounding boxes for candidate objects, then perform pixel-level segmentation within these boxes. Their strong generalization and segmentation capabilities have made them the current mainstream method for instance segmentation. Due to the vast diversity of zooplankton species requiring annotation by experienced biological experts, coupled with the necessity of specialized microscopy equipment to capture zooplankton microscopic images, no relevant instance segmentation datasets exist in the field of zooplankton research. Therefore, this paper establishes the first zooplankton microscopic image instance segmentation dataset to advance instance segmentation studies in this domain.

\subsection{Segment Anything Model (SAM)}
SAM\cite{SAM} is a prompt-engineering-based large-scale visual segmentation model pre-trained on 11 million images and over 1.1 billion masks. Its robust zero-shot segmentation capability enables transfer to various downstream visual tasks such as image restoration\cite{yu2023inpaint, 10657905} and style transfer\cite{Liu_2023}, as well as other vertical domains including medicine, remote sensing, and underwater applications\cite{SAM-Adapter, Hi-SAM, RSPrompter, USIS-SAM, ROS-SAM, FoodSAM, EviPrompt, SAMRS, 10418101}. For instance, FoodSAM\cite{FoodSAM} integrates SAM's robust segmentation capabilities with domain-specific semantic knowledge through semantic enhancement strategies, achieving semantic segmentation, instance segmentation, and panoramic segmentation of food images via object detectors. USIS-SAM\cite{USIS-SAM} fine-tunes SAM's encoder to learn underwater domain knowledge and generates salient features using a salient feature prompt generator, enabling instance segmentation of underwater images. However, SAM's lack of domain-specific knowledge for zooplankton and the distinct distribution of zooplankton microscopic images compared to other datasets limit its performance in segmenting such images. In this study, we leverage a zooplankton microscopic image instance segmentation dataset to build the first large-scale instance segmentation model for the zooplankton domain based on SAM.

\subsection{Wavelet Transform}
The Wavelet Transform adaptively adjusts the analysis scale by scaling and translating a mother wavelet function, enabling multi-scale and multi-directional signal analysis.  In recent years, several studies have integrated wavelet transform with deep learning to enhance model performance in image processing \cite{10196309, 1588962, Waveletadaptive} and intelligent recognition tasks \cite{wavemamba, WaveletConvolutions, BWG, 9933881, WaveSNet}. In the field of image segmentation, Xu \emph{et al.} \cite{XU2023109819} addressed the problem of information loss caused by conventional downsampling by introducing a Haar wavelet-based downsampling method. As a result, it reduces feature map resolution while preserving structural and textural details, demonstrating improved performance on public semantic segmentation datasets. For challenging segmentation tasks under low-light or adverse weather conditions such as rain and fog, Wang \emph{et al.} \cite{WaveCRNet} proposed a wavelet-constrained semantic segmentation framework. Their method incorporates the dual-tree complex wavelet transform within a multi-branch architecture and employs a pixel attention mechanism to fuse information from both the wavelet and feature domains, effectively improving segmentation accuracy in complex railway scenes. Unlike the aforementioned works, which do not introduce any learnable or weighted operations, our work focuses on dynamically fusing wavelet features for instance segmentation. We propose a wavelet-based multi-scale and multi-directional feature enhancement module designed to recover fine details lost, thereby improving boundary segmentation accuracy for transparent zooplankton in complex microscopic imaging conditions.

\section{Methodology}
In this section, we will introduce our proposed ZMIS-SAM model, a SAM-based framework that is designed specifically for instance segmentation of zooplankton microscopic images. The subsections are as follows: 3.1 Overall Framework of ZMIS-SAM, 3.2 Zooplankton Micrograph Adaptive Vision Transformer (ZM-ViT), 3.3 Neighboring Feature Aggregation Module (NFAM), and 3.4 Wavelet-based Multi-scale and Multi-directional Feature Enhancement (WM2FE). For more prior knowledge, the loss function and evaluation metrics of the model, please refer to \textbf{Appendix B.}

\subsection{Overall Framework of ZMIS-SAM}
\begin{figure*}[tb]
\centering
\includegraphics[width=4.5in]{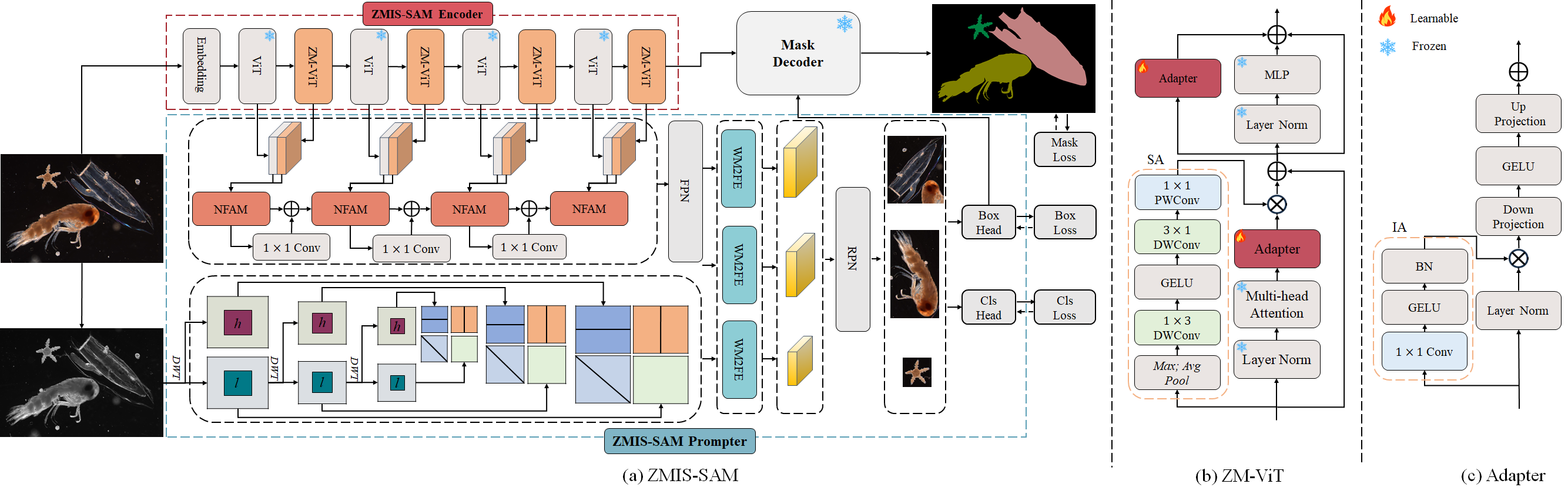}
\caption{Overview of ZMIS-SAM architecture. ZMIS-SAM includes ZMIS-SAM Encoder, ZMIS-SAM Prompter, and Mask Decoder.}
\label{fig: ZMIS-SAM}
\end{figure*}

The overall architecture of ZMIS-SAM is illustrated in \cref{fig: ZMIS-SAM}, which consists of three key components: ZM-ViT, NFAM, and WM2FE. Specifically, ZM-ViT incorporates two lightweight adapters, the Shape Adapter (SA) and the Intensity Adapter (IA). The SA refines the representation of zooplankton morphological structures, while the IA adjusts the channel-wise feature intensity distribution. Together, they guide SAM to capture domain-specific characteristics better and extract more discriminative zooplankton features. The NFAM aggregates the zooplankton-specific features extracted by ZM-ViT with the general visual features from adjacent frozen ViT layers, followed by a Feature Pyramid Network \cite{8099589} to obtain multi-scale representations. Furthermore, the WM2FE module performs wavelet decomposition on the multi-scale features to enhance both low-frequency contextual cues and high-frequency directional details, effectively compensating for the edge information lost during the encoding process. Finally, the features from the encoder and the embeddings from the prompt encoder are fed into the frozen mask decoder of SAM to produce fine-grained zooplankton instance segmentation results.

\subsection{ZM-ViT: Zooplankton Micrograph Adaptive ViT}
The SAM's image encoder can extract rich and general-purpose visual representations that are effective across many vertical domains. However, zooplankton in marine environments often exhibit slender appendages, transparent or semi-transparent bodies, and significant illumination variations caused by the microscopic imaging environment. Moreover, there exists a notable domain gap between the natural images used for SAM pretraining and the microscopic zooplankton images. These factors make the original SAM image encoder less effective in capturing the fine-grained morphological characteristics of zooplankton, leading to suboptimal segmentation performance. To address this issue, we propose ZM-ViT, which integrates two lightweight adapters into the ViT blocks of SAM, as shown in \cref{fig: ZMIS-SAM}(b). This design enables ZM-ViT to effectively leverage the pretrained SAM image encoder while adapting to the zooplankton domain, allowing for more discriminative feature extraction and improved instance segmentation accuracy.

The proposed ZM-ViT can effectively learn discriminative features of zooplankton in microscopic imagery through Parameter-Efficient Fine-Tuning \cite{Wang2024ParameterEfficientFI}. Inspired by USIS-SAM, we integrate two lightweight adapters into the ViT architecture, namely the Intensity Adapter (IA) and the Shape Adapter (SA). Unlike the adapter design in USIS-SAM, we introduce a parallel branch that combines Conv–GELU–BN operations with the Layer Normalization (LN) residual pathway for the features fed into the adapter. This design forms the IA, which dynamically adjusts the channel-wise feature intensity distribution. The placement of IA within the ViT block remains consistent with that of the adapter in USIS-SAM: one IA module is inserted after the multi-head self-attention layer to modulate the channel intensity distribution, and another is integrated in parallel with the frozen MLP residual branch to aggregate and update contextual information more effectively. The IA can be formulated as follows:
\begin{equation}
\begin{aligned}
  F_{IA} = \mathcal{U}_{p}(\mathcal{D}_{p}(BN(LN(x) * \mathcal{C}(x)))) + {F}_{in}
\end{aligned}
\end{equation}
where $x$ and $F_{IA} \in \mathbb{R}^{N\times C}$ denote the input and output features, respectively. $\mathcal{C}(\cdot)$ denotes the convolutional branch consisting of Conv–GELU–BN, $\mathcal{U}_{p}(\cdot)$ and $\mathcal{D}_{p}(\cdot)$ denote the up projection and down projection, respectively. $BN(\cdot)$ and $LN(\cdot)$ denote the batch normalization and layer normalization, respectively.

Since zooplankton have diverse morphological structures, we further design a Shape Adapter (SA) based on a striped depth-wise separable convolution. The SA can effectively enhance the model’s ability to represent the shape structural characteristics of zooplankton, while the depthwise separable convolution greatly reduces computational complexity. The SA can be expressed as follows:
\begin{equation}
\begin{aligned}
  F_{SA} =  \sum_{max;avg} \mathcal{C}_{1}((Pool_{i}(x)))
\end{aligned}
\end{equation}
where $x$ and $F_{SA} \in \mathbb{R}^{N\times C}$ denote the input and output features, respectively.  $\mathcal{C}_{1}(\cdot)$ denotes the PWConv-DWConv-GELU-DWConv operation. $Pool(\cdot)$ denotes average pooling or maximum pooling.

To enable the SAM image encoder to extract optimal feature representations, we use ViT-H as the backbone of the SAM image encoder. Furthermore, to ensure a fair comparison with the baseline experiments, ZM-ViT replaces one frozen ViT layer every two layers, starting from the eighth layer.

\subsection{NFAM: Neighboring Feature Aggregation Module}
To achieve automatic instance segmentation, existing approaches usually treat the representations extracted from certain ViT layers of the SAM image encoder as pseudo-prompts. However, when transferring SAM to domain-specific instance segmentation tasks, current methods either rely solely on the frozen ViT features with general representations or only exploit the domain-specific representations obtained from the adapter. Both strategies fail to fully leverage the complementary information between general and domain-specific representations for generating effective prompt features. 

\begin{figure}[tb]
\centering
\includegraphics[width=3.0in]{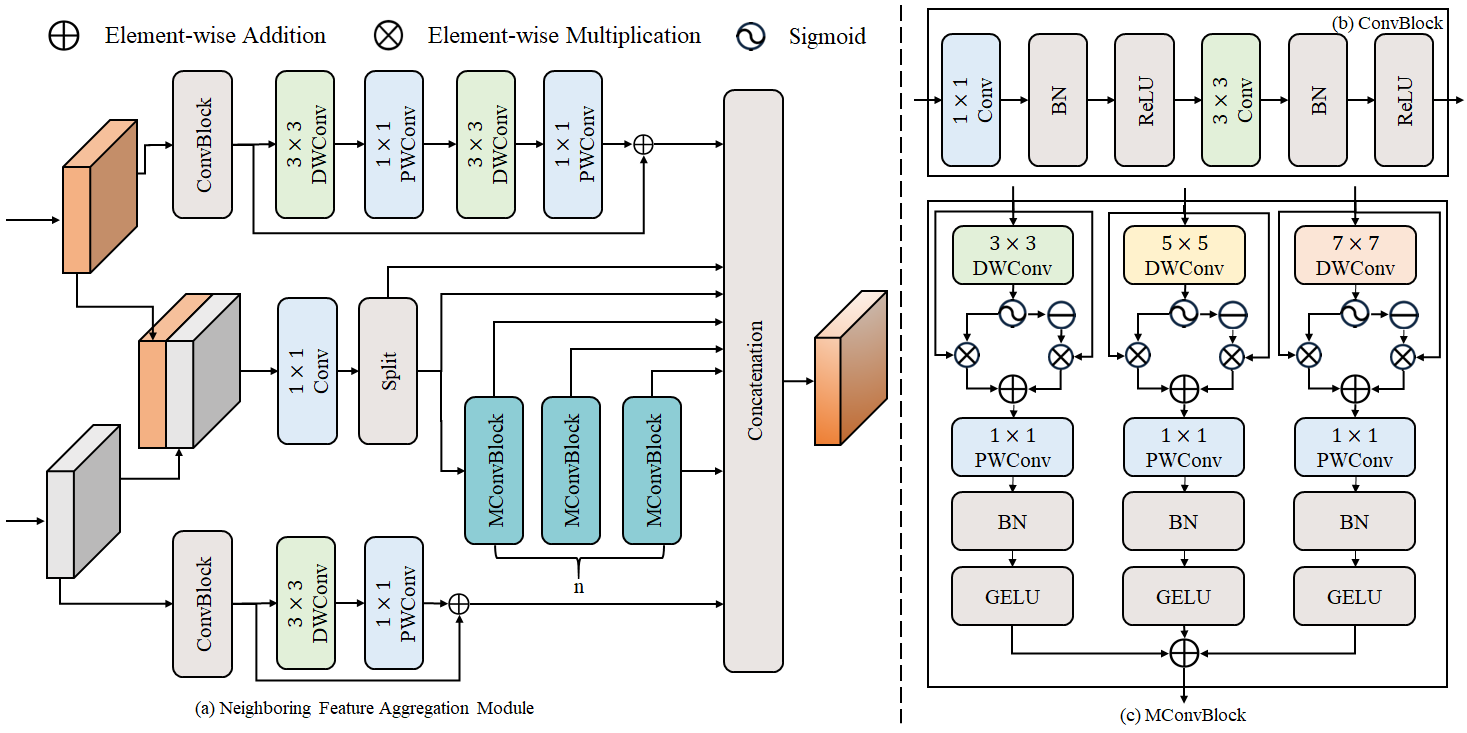}
\caption{Overview of Neighboring Feature Aggregation Module (NFAM) architecture.}
\label{fig: NFAM}
\end{figure}

We propose an NFAM that comprehensively aggregates both the general representations extracted from the frozen ViT layers and the domain-specific features captured by the fine-tuned ZM-ViT, thereby producing more effective prompt representations and enhancing the instance segmentation performance of SAM. As illustrated in \cref{fig: NFAM}, NFAM processes the domain-specific features $\{F_{sdf}\}_{i=j}^{L} \in \mathbb{R}^{C_{i}\times H_{i}\times W_{i}}$, the general features $\{F_{gf}\}_{i=j-1}^{L-1} \in \mathbb{R}^{C_{i}\times H_{i}\times W_{i}}$, and their concatenated representations $F_{cf} \in \mathbb{R}^{2C_{i}\times H_{i}\times W_{i}}$ in parallel. The domain-specific and general features are respectively refined using a ConvBlock (\cref{fig: NFAM} (b)) and a depthwise separable convolution to enhance local feature extraction, while residual connections preserve fine-grained spatial information. 

\begin{equation}
\begin{aligned}
  \tilde{F_{sdf}} = ConvBlock(F_{sdf}) +\mathcal{C}_{2}(ConvBlock(F_{sdf}))
\end{aligned}
\end{equation}
\begin{equation}
\begin{aligned}
  \tilde{F_{gf}} = ConvBlock(F_{gf}) + \mathcal{C}_{3}(ConvBlock(F_{gf}))
\end{aligned}
\end{equation}
\begin{equation}
\begin{aligned}
  F_{cf} = Concat(F_{sdf},F_{gf})
\end{aligned}
\end{equation}
where $\tilde{F_{sdf}}$ and $\tilde{F_{gf}}$ domain-specific and general features after local enhancement, respectively. $\mathcal{C}_{2}(\cdot)$ and  $\mathcal{C}_{3}(\cdot)$ denote the PWConv-DWConv-PWConv-DWConv and the PWConv-DWConv operation, respectively. In parallel, the concatenated features are passed through a $1\times1$ convolution to balance the channel dimensions. The resulting feature map is then divided into $n$ parts, each processed by an independent multi-scale convolutional block (\cref{fig: NFAM} (c)) with different receptive fields to capture spatial details at multiple scales. 
\begin{equation}
\begin{aligned}
  \lbrack F^{(1)},...,F^{(n)} \rbrack = Split_{n}(Conv_{1\times 1}(F_{cf}))
\end{aligned}
\end{equation}
\begin{equation}
\begin{aligned}
  {M^{(j)}} = MConvBlock(F^{(j)}), j=1,...,n
\end{aligned}
\end{equation}
where $Split_{n}(\cdot)$ denotes dividing the input representation into $n$ parts. $F^{(j)}$ denotes the $j$-th characteristic obtained from the division. $M^{(j)}$ denotes the $j$-th part after multi-scale convolution processing. 

In the MConvBlock, depth-wise convolutions with different receptive fields are first used to extract multi-scale features from the input features separately. Then, a gating mechanism is employed to dynamically adjust the weights of features at different scales. Pointwise convolution, Batch Normalization, and GELU activation follow each receptive field is utilized to output the enhanced representation. 
\begin{equation}
\begin{aligned}
  \psi_{k} = \sigma(DW_{k\times k}(x))
\end{aligned}
\end{equation}
\begin{equation}
\begin{aligned}
  x_{k}^{gate} = ( \psi_{k} * x + (1- \psi_{k}) * x))
\end{aligned}
\end{equation}
\begin{equation}
\begin{aligned}
  B_{k}(x) = GELU(BN(PW_{1\times 1}(x_{k}^{gate}))
\end{aligned}
\end{equation}
\begin{equation}
\begin{aligned}
  MConvBlock(x) = \sum_{k=3}^{K} B_{k}(x), k=3,5,7
\end{aligned}
\end{equation}
where $\psi_{k}$ denotes the gating attention weights under different receptive fields. $\sigma(\cdot)$ and $GELU(\cdot)$ denote the activation function of the sigmoid and GELU, respectively. $x_{k}^{gate}$ denotes the gating characteristic. $BN(\cdot)$ denotes the Batch Normalization. $B_{k}(x)$ refers output feature of the receptive field $k$ branch. Finally, the outputs from all branches are concatenated to form the aggregated representation $ F_{a}$.
\begin{equation}
\begin{aligned}
  F_{a} = Concat(M^{(1)},M^{(2)},...,M^{(n)}, \tilde{F_{sdf}}, \tilde{F_{gf}})
\end{aligned}
\end{equation}

\subsection{WM2FE: Wavelet-based Multi-scale and Multi-directional Feature Enhancement}
\begin{figure}[tb]
\centering
\includegraphics[width=3.0in]{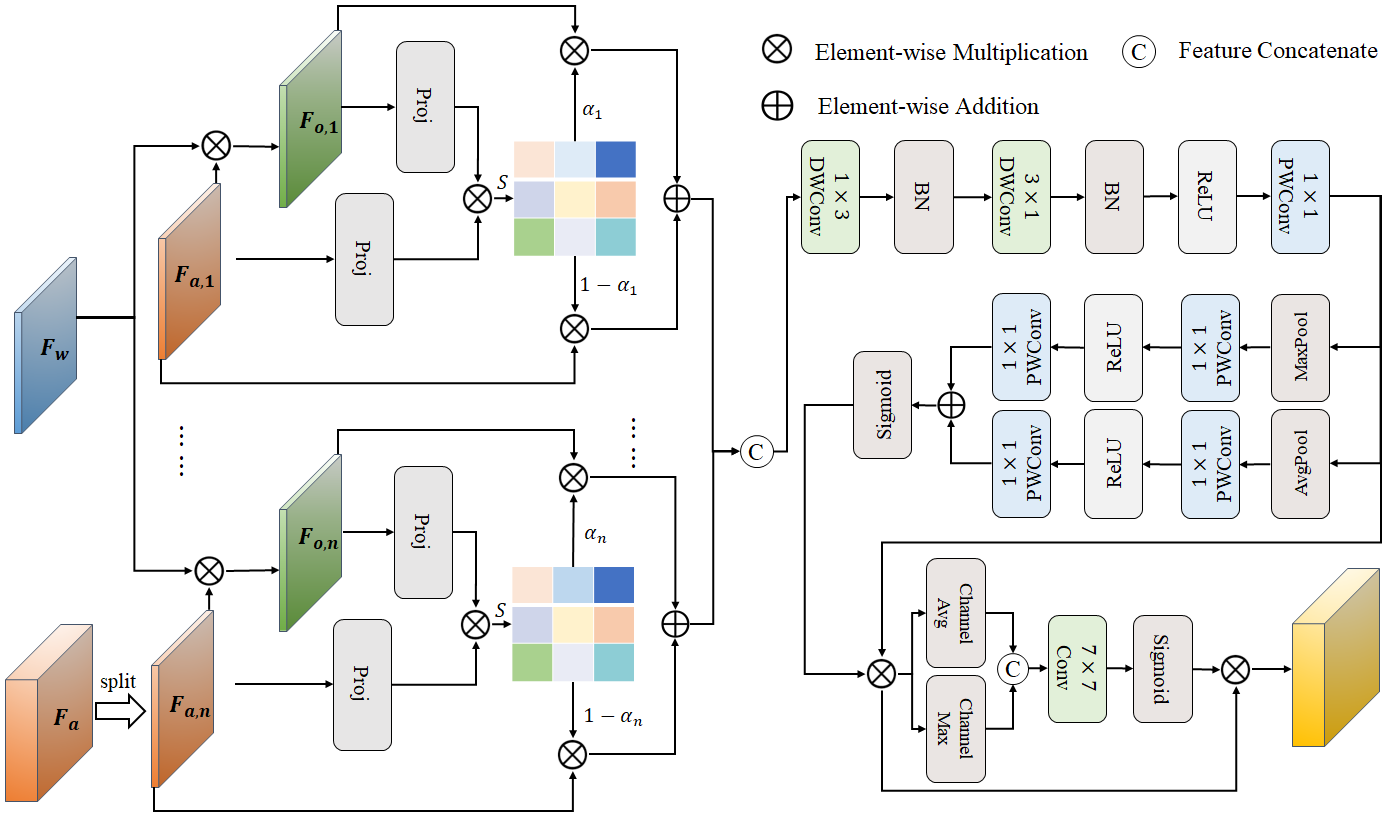}
\caption{Overview of Wavelet-based Multi-scale and Multi-directional Feature Enhancement (WM2FE) architecture. }
\label{fig: WM2FE}
\end{figure}

After obtaining the aggregated features, existing SAM-based instance segmentation methods usually employ a simple FPN network to extract multi-scale features and generate candidate prompts for enhancing model robustness. However, in the FPN, upsampling low-resolution features to high-resolution ones via deconvolution may lead to the loss of fine-grained details. As a multi-scale and multi-directional image analysis tool, the wavelet transform can provide high-frequency details in vertical, horizontal, and diagonal directions. Therefore, we design a Wavelet-based Multi-scale Feature Enhancement (WM2FE) module to compensate for the missing details during upsampling and improve the instance segmentation performance for zooplankton. 

As illustrated in \cref{fig: WM2FE}, WM2FE enhances multi-scale feature representation through dynamic fusion of aggregated features and wavelet-weighted features. Specifically, the aggregated feature $F_{a}$ is divided into $n$ parts $\{ F_{a,1}, F_{a,2},...F_{a,n} \}$ along the wavelet feature $F_{w}$ dimension, and each part is modulated by the corresponding wavelet feature to strengthen the representation of transparent boundaries of zooplankton. Then, linear learnable parameters are applied to dynamically fuse the aggregated features and wavelet-enhanced features. Finally, the $n$ parts are concatenated along the channel dimension, and a directional depthwise separable convolution is employed to enhance directional perception.

\begin{equation}
\begin{aligned}
   F_{o,i} =  F_{a,i}\odot F_{w}
\end{aligned}
\end{equation}
\begin{equation}
\begin{aligned}
  F_{m,i} = \alpha_{i} F_{o,i} + (1-\alpha_{i}) F_{a,i}, i=1,2,...,n
\end{aligned}
\end{equation}
\begin{equation}
\begin{aligned}
  F_{m} = \mathcal{C}_{4}(Concat(F_{m,1}, F_{m,2}, ..., F_{m,n}))
\end{aligned}
\end{equation}
where $\mathcal{C}_{4}(\cdot)$ denotes the PWConv-ReLu-DWConv-BN-DWConv operation. $\odot$ denotes element-wise multiplication, $\alpha_{i}$ denotes the linear learnable parameter of the $i$-th part. $F_{w}=Contact[LL; LH; HL; HH]$ denotes the wavelet features.

To further focus on informative regions, inspired by \cite{CBAM}, we introduce channel and spatial attention mechanisms in WM2FE. The channel attention module learns channel-wise weights by feeding both average-pooled and max-pooled features into a shared MLP followed by a sigmoid activation function.
\begin{equation}
\begin{aligned}
  M_{c} = \sigma(\sum_{avg;max} \mathcal{C}_{5}(F_m))
\end{aligned}
\end{equation}
\begin{equation}
\begin{aligned}
  F^{'} = M_{c} \otimes F_m
\end{aligned}
\end{equation}
where $F^{'}$ denotes the refined features. $M_{c}$ denotes the channel attention map and $\sigma$ denotes the sigmoid function. $\mathcal{C}_{5}(\cdot)$ denotes the PWConv-ReLu-DWConv-BN-DWConv-BN-DWConv operation. 

Subsequently, the spatial attention mechanism aggregates average-pooled and max-pooled features along the channel dimension, followed by a 7×7 convolution and sigmoid activation to generate the spatial attention map:
\begin{equation}
\begin{aligned}
  M_{s} = \sigma(f^{7\times 7}([\mathcal{A}(F^{'}); \mathcal{M}(F^{'})]))
\end{aligned}
\end{equation}
\begin{equation}
\begin{aligned}
  F_{fe} = M_{s} \otimes F^{'}
\end{aligned}
\end{equation}

\section{Experiments}
In this section, we evaluate the performance of ZMIS-SAM on the ZMIS5K dataset and compare it with SOTA instance segmentation models, and further conduct ablation experiments. In addition, we conducted generalization experiments on public datasets from other domains, as detailed in \textbf{Appendix D.}

\subsection{Datasets}
The ZMIS5K dataset contains a total of 5,358 images, including 47 species of zooplankton, and was divided into 4,262 images for the training set and 1,096 images for the test set. For a more detailed introduction to the ZMIS5K dataset, please refer to \textbf{Appendix A.}

\subsection{Implementation details}
We build our ZMIS-SAM model based on the MMDetection framework \cite{mmdetection}, and most of the comparison models in this paper are also trained and evaluated within the same framework. ZMIS-SAM and other SAM-based comparison models are trained on 6 NVIDIA RTX 3090 GPUs, with a batch size of 2 per GPU, for 400 epochs. All input images are resized to 512 × 512 pixels, and standard data augmentation techniques, including random flipping, random scaling, and random cropping, are applied to enhance the model’s generalization capability. We use the AdamW optimizer and adopt automatic mixed precision to accelerate training and reduce GPU memory consumption. The initial learning rate is set to 2.0e-4. In the first stage, a linear warm-up is performed over the first 50 iterations, followed by a cosine decay schedule to gradually reduce the learning rate in the second stage. 
\begin{table*}
\centering
\caption{\label{tab: result}A comparison with SOTA methods on ZMIS5K is provided. Results highlighted in red indicate the best performance, while those in blue denote the second-best.}
\begin{tabular}{lcc|ccc}
\hline
Method & Backbone & Params & $mAP$ & $AP_{50}$ & $AP_{75}$ \\\hline
Mask R-CNN \emph{(ICCV'17)}\cite{MaskRCNN} & ResNet-101 & 63M & 66.4 & 90.9 & 69.0  \\
Cascade Mask R-CNN \emph{(CVPR'18)}\cite{cai18cascadercnn} & ResNet-101 & 88M & 66.0 & 89.5 & 68.3 \\
Mask Scoring R-CNN \emph{(CVPR'19)}\cite{8953609} & ResNet-101 & 79M & 67.0 & 89.8 & 69.1 \\
CondInst \emph{(ECCV'20)}\cite{tian2020conditional} & ResNet-101 & 63M & 68.3 & 91.7 & 70.9 \\
$R^{3}$-CNN \emph{(CAIP'21)}\cite{R3} & ResNet-101 & 77M & 66.6 & 90.0 & 70.0 \\
ConvNeXt \emph{(CVPR'22)}\cite{liu2022convnet} & ConvNeXt-T & 63M & 68.4 & 92.4 & 70.3 \\
Mask2Former \emph{(CVPR'22)}\cite{Mask2former} & ResNet-101 & 63M &70.7 & 87.4 & 76.1 \\
WaterMask \emph{(ICCV'23)}\cite{WaterMask} & ResNet-101 & 67M & 60.7 & 82.2 & 64.4\\\hline
ViT-UWA \emph{(TIP'26)}\cite{jia2026vit} & ViT-B & 129M & 62.3 & 90.7 & 61.7 \\
SAM+Bbox \emph{(ICCV'23)}\cite{SAM} & ViT-H & 641M & 60.2 & 82.9 & 65.2 \\
SAM+Mask \emph{(ICCV'23)}\cite{SAM} & ViT-H & 641M & 70.1 & \textcolor{blue}{\textbf{94.0}} & 75.5 \\
RSPrompter \emph{(TGRS'24})\cite{RSPrompter} & ViT-H & 632M & \textcolor{blue}{\textbf{71.8}} & 92.2 & \textcolor{blue}{\textbf{76.9}} \\
USIS-SAM \emph{(ICML'24)}\cite{USIS-SAM} & ViT-H & 700M & 70.4 & 91.7 & 75.8 \\
UCIS-SAM \emph{(TIP'26)}\cite{11455610} & ViT-H & 723M & 66.9 & 80.4 & 72.1 \\
ZMIS-SAM & ViT-H & 758M & \textcolor{red}{\textbf{73.6}} & \textcolor{red}{\textbf{94.6}} & \textcolor{red}{\textbf{80.7}}\\\hline
\end{tabular}
\end{table*}

\subsection{Experiments Result}

\textbf{Quantitative Results.} We compare our ZMIS-SAM with fourteen state-of-the-art instance segmentation models on the ZMIS5K test set. To explore the optimal segmentation performance of each model, traditional instance segmentation methods are implemented with ResNet-101 as the backbone, while SAM-based methods adopt ViT-H, the largest variant of the SAM backbone, to ensure fair comparison. As shown in \cref{tab: result}, our proposed ZMIS-SAM achieves the best overall performance across all instance segmentation metrics. Traditional methods such as Mask R-CNN, CondInst, and ConvNeXt exhibit limited accuracy, with $mAP$ values ranging from 66.0\% to 68.4\%. Among traditional models, Mask2Former achieves the highest 70.7\%, 87.4\%, and 76.1\% AP in $mAP$, $AP_{50}$ and $AP_{75}$, however, ZMIS-SAM surpasses it by 2.9\% in $mAP$, 7.2\% in $AP_{50}$, and 4.6\% in $AP_{75}$, demonstrating the superior capability of our model in capturing fine-grained object boundaries. Compared with traditional models, SAM-based instance segmentation approaches achieve better performance owing to their strong feature extraction and prompt-based representation learning. Nevertheless, our ZMIS-SAM further improves segmentation accuracy with only a small increase in model parameters, outperforming the best SAM-based model RSPrompter by 1.8\% in $mAP$, 2.4\% in $AP_{50}$, and 3.8\% in $AP_{75}$.
\begin{figure}[tb]
\centering
\includegraphics[width=4.0in]{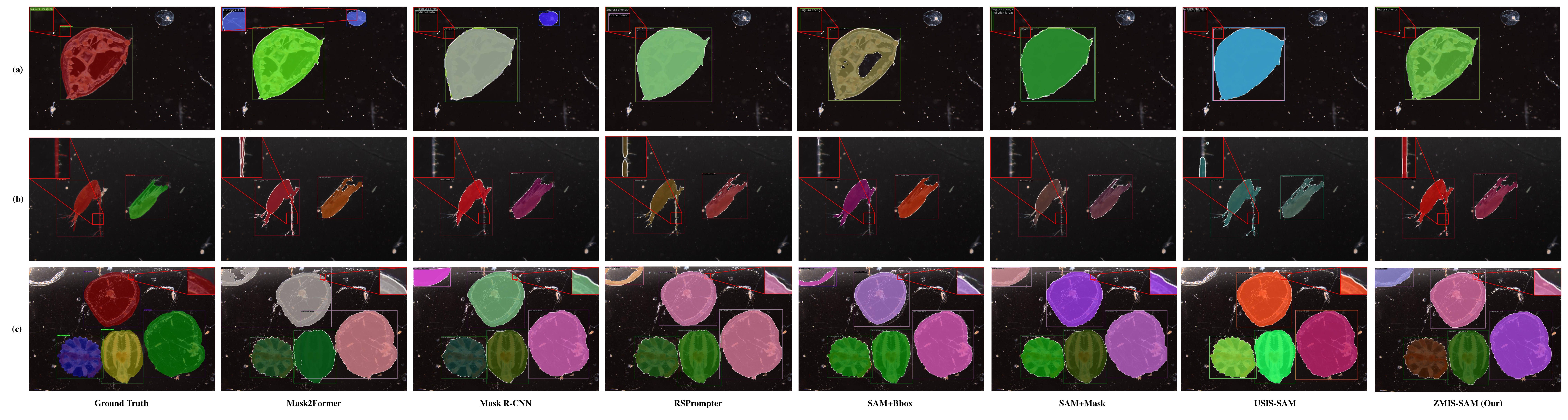}
\caption{Quantitative comparison between the ZMIS-SAM model and other state-of-the-art models on the ZMIS5K dataset. (a), (b), and (c) denote inaccurate classification, discontinuous segmentation of slender appendages, and incomplete boundary segmentation, respectively. Higher-resolution visualizations can be found in the appendix.}
\label{fig: Result}
\end{figure}

\textbf{Qualitative Results.} We further conduct qualitative comparisons between ZMIS-SAM and several top-performing instance segmentation models on the ZMIS5K test set. To investigate the limitations of SAM when transferred to vertical domains, we visualize the segmentation results and analyze them in detail. As shown in \cref{fig: Result}, the proposed ZMIS-SAM not only distinguishes different zooplankton species accurately but also captures distinctive morphological characteristics such as slender appendages, while performing fine-grained boundary segmentation. Specifically, in \cref{fig: Result} (a), our method successfully differentiates between \emph{Jellyfish larvae} and \emph{Sugiura chengshanense}, while avoiding false segmentation of unannotated organisms. In contrast, Mask2Former incorrectly segments unidentified organisms (e.g., \emph{Fish eggs}) as known species. In \cref{fig: Result} (b), ZMIS-SAM effectively captures the relationship between the main body and slender appendages of \emph{Calanus sinicus}, achieving complete and coherent segmentation. Although RSPrompter and USIS-SAM can also segment the appendages, their results are fragmented and discontinuous, unlike our method’s continuous representation. In \cref{fig: Result} (c), ZMIS-SAM demonstrates superior boundary refinement ability. For example, Mask R-CNN fails to delineate the boundaries of \emph{Pleurobrachia globosa}, while our approach achieves precise and consistent edge segmentation. These qualitative comparisons further confirm that ZMIS-SAM excels in recognizing fine morphological details, maintaining boundary integrity, and suppressing false detections, addressing key challenges in SAM’s transfer to specialized domains such as microscopic zooplankton segmentation.

\subsection{Ablation experiments on ZMIS-SAM}
In this section, we conduct a series of ablation studies on the ZMIS5K dataset to evaluate the design of ZMIS-SAM systematically. The experiments are designed to investigate: 1) the effectiveness of individual modules and 2) the influence of different wavelet basis functions on the final segmentation results. More ablation experiments and parametric analysis are provided in \textbf{Appendix C.}
\begin{minipage}[c]{0.6\textwidth}
    \centering
    \captionof{table}{\centering Albation study of our contributions.}
    \label{tab: ablation-result1}
    \setlength{\tabcolsep}{0.5 mm}
    \scalebox{0.9}{
    \begin{tabular}{cccccc}
        \hline
        ZM-ViT & NFAM & WM2FE & $mAP$ & $AP_{50}$ & $AP_{75}$ \\\hline
        &  &  & 70.4 & 91.7 & 75.8  \\
        \textbf{$\checkmark$} &  &  & 71.7 & 92.0 & 77.7 \\
        \textbf{$\checkmark$} & \textbf{$\checkmark$} &  & 73.0 & 94.4 & 78.7 \\
        \textbf{$\checkmark$} &  & \textbf{$\checkmark$} & 72.7 & 94.1 & 79.4 \\
        \textbf{$\checkmark$} & \textbf{$\checkmark$} & \textbf{$\checkmark$} & \textbf{73.6} & \textbf{94.6} & \textbf{80.7}\\\hline
    \end{tabular}
    }
\end{minipage}
\hfill
\begin{minipage}[c]{0.4\textwidth}
    \captionof{table}{\centering Albation study of wavelet function for the WM2FE.}
    \label{tab: ablation-wt}
    \centering
    \setlength{\tabcolsep}{0.5 mm}
    \scalebox{0.9}{
    \begin{tabular}{lccc}
        \hline
        $\varphi$ & $mAP$ & $AP_{50}$ & $AP_{75}$ \\\hline
        rbio1.1   & 72.7 & 93.8     & 79.2 \\
        bior1.1   & 72.0 & 92.6 & 78.4\\
        haar   & \textbf{73.6} & \textbf{94.6} & \textbf{80.7} \\\hline
    \end{tabular}
    } 
\end{minipage}

\textbf{Effectiveness of Individual Modules.} We perform an incremental ablation study to systematically evaluate the individual contribution and combined effect of each proposed module for zooplankton instance segmentation in microscopy images. The complete results are presented in \cref{tab: ablation-result1}, where we sequentially evaluate the following configurations: 1) whether to use ZM-ViT, 2) whether to incorporate the NFAM, and 3) whether to employ the WM2FE module. The results demonstrate that each module consistently improves instance segmentation performance. Using USIS-SAM with a ViT-H backbone as the baseline, we observe that replacing its original ViT with our ZM-ViT brings a 1.3\% $mAP$ gain with only a marginal increase in parameters. Building upon ZM-ViT, the further addition of NFAM and WM2FE improves $mAP$ by 1.3\% and 1.0\%, respectively. When both NFAM and WM2FE are integrated, ZMIS-SAM achieves $mAP$, $AP_{50}$, and $AP_{75}$ scores of 73.6\%, 94.6\%, and 80.7\%, respectively, establishing a new state-of-the-art in zooplankton instance segmentation with only a modest parameter overhead.

\textbf{Optimal Wavelet Function for the WM2FE.} We evaluate the impact of different wavelet basis functions on the instance segmentation performance of ZMIS-SAM, with experimental results summarized in \cref{tab: ablation-wt}. Different wavelets exhibit distinct frequency domain response characteristics when extracting image features, which influences the model's ability to capture fine object details. Experimental results indicate that using the Haar wavelet in the WM2FE module yields the best performance across $mAP$, $AP_{50}$, and $AP_{75}$, achieving scores of 73.6\%, 94.6\%, and 80.7\%, respectively, surpassing both rbio1.1 and bior1.1 wavelets. We attribute this improvement to the high sensitivity of the Haar wavelet to edge structures, which enables more effective capture of zooplankton contour and boundary information, thereby enhancing the instance segmentation capability of ZMIS-SAM.

\section{Limitations}
Although the ZMIS-SAM model can perform instance segmentation on microscopic images of zooplankton, differences in task definition prevent the framework from separately segmenting the heads and antennae of zooplankton. In future work, we will introduce taxonomic and morphological text information to construct a multimodal framework for the zooplankton domain to achieve fine‑grained segmentation of zooplankton body structures.

\section{Conclusion}
In this study, we introduce the first high-quality zooplankton microscopy image segmentation dataset (ZMIS5K), spanning 47 species with 5,358 images and 10,228 instances. More importantly, we propose ZMIS-SAM, the first SAM-based instance segmentation model specifically designed for zooplankton microscopy images. In ZMIS-SAM, we innovatively propose three modules: ZM-ViT, NFAM, and WM2FE, which effectively address the issues of inaccurate classification, discontinuous segmentation of slender appendages, and incomplete boundary segmentation encountered when transferring SAM to the zooplankton domain. Extensive experiments demonstrate that ZMIS-SAM achieves state-of-the-art performance on the ZMIS5K dataset and exhibits strong generalization capability on cross-domain benchmarks.

\section*{Acknowledgements}
This work is supported in part by the Natural Science Foundation of Shandong Province (Grant No. ZR2024MF037). The North China Sea Ecological Center of the Ministry of Natural Resources strongly supports this work. The center provided samples for zooplankton photography and the microscopic imaging equipment required for this study. In addition, teachers, including Xiaoli Song and Yanping Qi, offered detailed guidance on zooplankton biology.

%
%
\bibliographystyle{splncs04}
\bibliography{main}
\appendix
\onecolumn
\section{ZMIS5K Dataset}
\subsection{Dataset Collection and Annotation}
\begin{figure}
\centering
\includegraphics[width=4.5in]{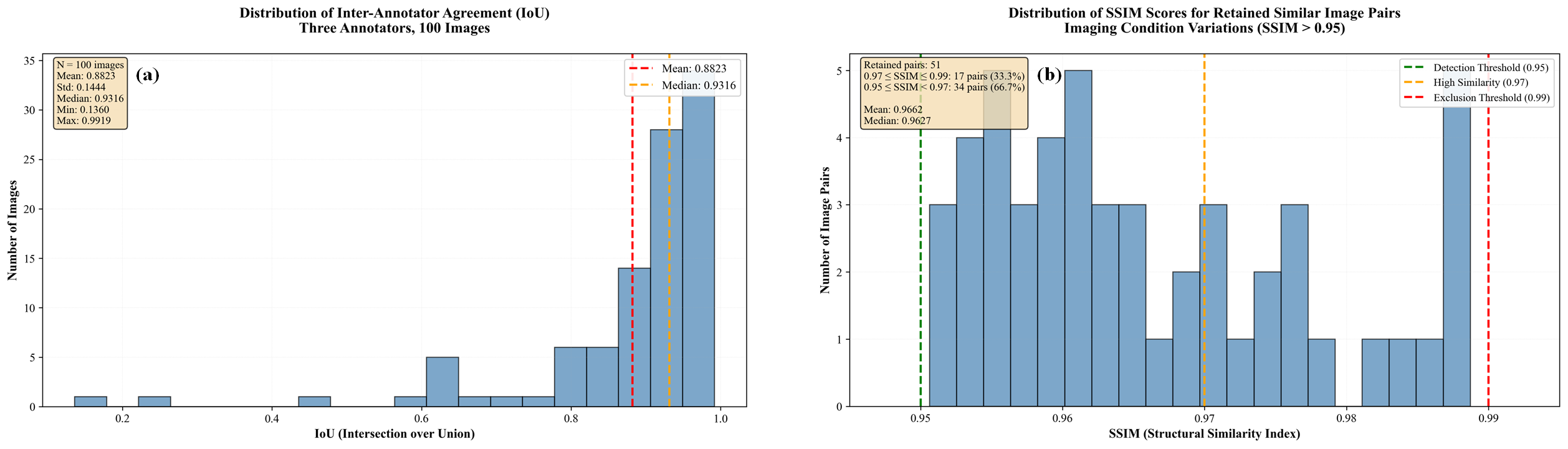}
\caption{Evaluation results of dataset construction quality.}
\label{fig: DA}
\end{figure}

\begin{figure}
\centering
\includegraphics[width=4.5in]{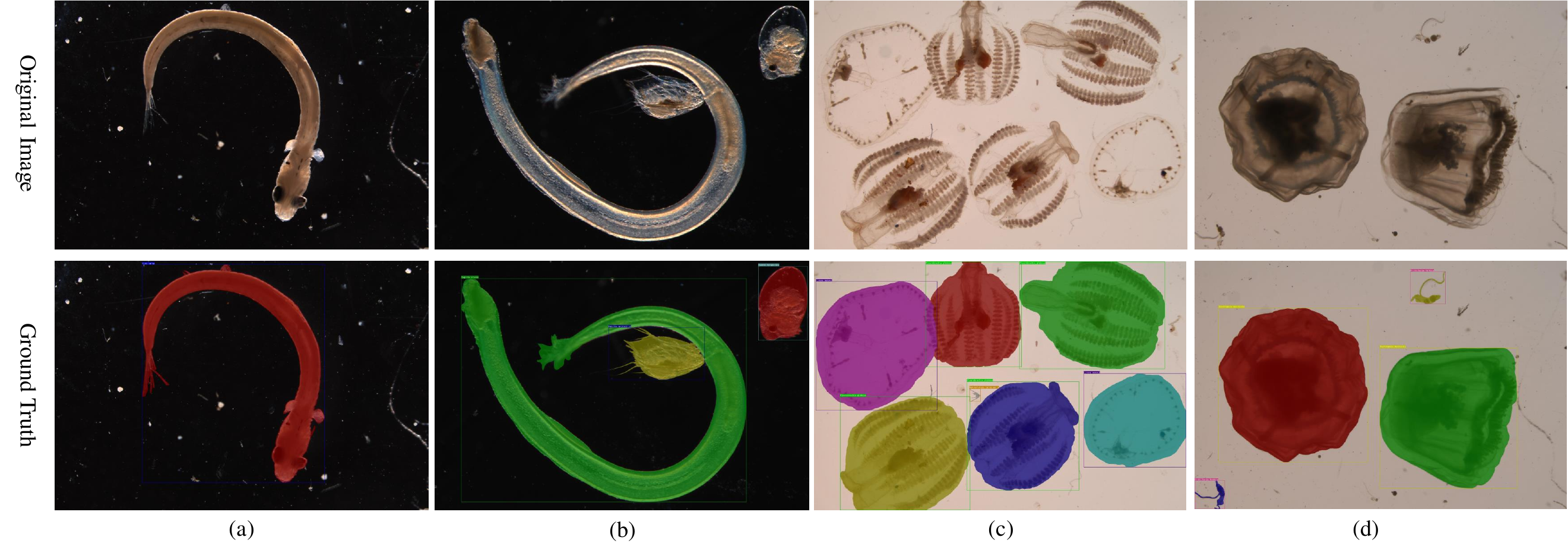}
\caption{Examples of annotations for various instances in ZMIS5K. (a) is \emph{Fish larva}. (b) includes \emph{Sagitta crassa}, \emph{Penilia avirostris}, and \emph{Evadne tergestina}. (c) includes \emph{Eirene menoni} and \emph{Pleurobrachia globosa}. (d) includes \emph{Turritopsis nutricula} and \emph{Oikopleura longicauda}.}
\label{fig: Categories}
\end{figure}

We captured 5830 microscopic images of zooplankton using a research-grade stereomicroscope, Leica M205A. This microscope has a magnification ratio of 20.5:1, can achieve a maximum magnification of 1280x based on a combination of optical components, and has a resolution of 1050 lp/mm. These images contain most species of zooplankton in the coastal waters of Shandong Province, such as \emph{Calanus sinicus}, \emph{Sagitta crassa}, and \emph{Macruran larvae}. Then, we perform pixel-level and category-level annotations on each zooplankton instance in these images. To ensure the quality and accuracy of data annotation as much as possible, we first have biological experts in the field of zooplankton perform category annotation on the captured images. Then, 10 annotators received training from the experts proficient in zooplankton biology. Annotations were independently reviewed and refined by three additional senior annotators. Quantitative evaluation (\cref{fig: DA}(a)) achieves a high pairwise IoU of 0.8823, confirming reliable annotation consistency. After removing unclear images and images with excessive impurities, we finally obtained 5358 images, which constitute ZMIS5K. The actual annotation results are shown in \cref{fig: Categories}.

\subsection{Dataset Characteristics and Statistics}
\begin{figure}
\centering
\includegraphics[width=4.0 in]{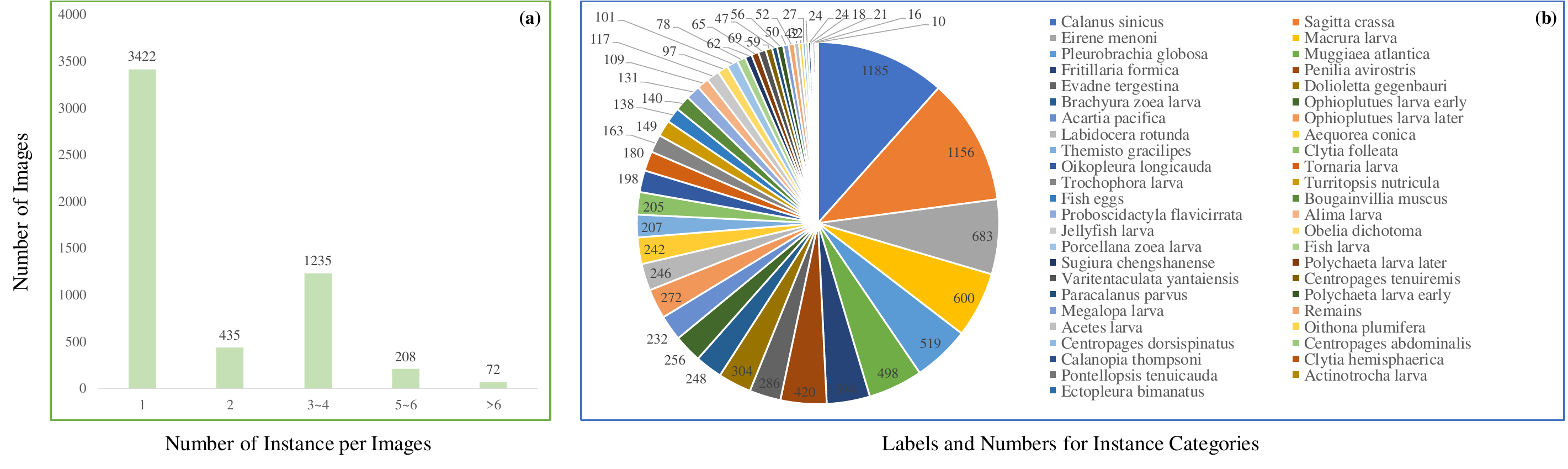}
\caption{Essential characteristics of the ZMIS5K dataset. (a) Distribution of the number of instances per image in the ZMIS5K dataset. (b) The number of instances per category in the ZMIS5K dataset.}\label{fig: Number}
\end{figure}

In this subsection, we will introduce the basic information, characteristics of ZMIS5K, and the challenges encountered in instance segmentation. Some statistical information can also be seen in \cref{fig: Number}.

\textbf{Number and Category of Dataset.} Under a microscope, different lighting conditions and magnification levels result in zooplankton exhibiting other characteristics, while also encompassing a wide range of categories. It is therefore necessary to enhance the model's generalization ability in complex scenarios and avoid overfitting. Therefore, the 5358 images were collected, each with a resolution of 2736×1824 pixels. Each species was imaged under two different illumination conditions, bright-field and dark-field, and captured at multiple magnification levels. The dataset contains 47 species of zooplankton and 1 background class, and these data are divided into a training set and a test set in an 8:2 ratio. In dataset partitioning, we adopted SSIM to check duplicate or near-identical samples between training and test sets. Samples with SSIM above 0.95 (\cref{fig: DA}(b)), caused merely by slight variations in illumination and focal plane during imaging, which conforms to practical real-world imaging conditions. Furthermore, to make the ZMIS5K dataset suitable for more downstream tasks, all annotated data include category labels, instance masks, and bounding boxes. This enables the ZMIS5K dataset to be used for semantic segmentation, instance segmentation, object detection, and other downstream tasks. 

\begin{table}
\centering
\caption{\label{tab: comparison}Comparison with existing plankton datasets. Where OD indicates object detection, SS indicates semantic segmentation, and IS indicates instance segmentation. Max indicates the maximum number of instances in a single image. $\geq3$ indicates the number of images where the number of instances contained in a single image is greater than or equal to 3.}
\begin{tabular}{lcccccccc}
\hline
Dataset & OD & SS & IS & Number & Categories & Max & $\geq3$ & Year\\\hline
PMID2019 \cite{PMID2019}  & $\checkmark$ & $\times$ & $\times$  & 10819 & 24  & 6 & 436 & 2019 \\
Bergum \emph{et.al}\cite{OCEANS}  & $\checkmark$ & $\checkmark$ & $\checkmark$ & 131 & 7  & 9 & 88 & 2020\\
ZMIS5K & $\checkmark$ & $\checkmark$ & $\checkmark$ & 5358 & 47  & 26 & 1497 & 2026\\\hline
\end{tabular}
\end{table}

\textbf{Number and Size of Instances.} There are many individual images containing multiple instances in the ZMIS5K dataset. In the dataset we constructed, the total number of instances is 10,228, among which the number of images with more than 3 instances amounts to 1,515 in total, accounting for 28.2\% of the total dataset size. In the phytoplankton object detection dataset, PMID2019 \cite{PMID2019}, it contains 24 species of phytoplankton, with a total of 10,819 images and 13,817 instances in total, where images with more than three instances account for 4.0\% of the total. In the in-situ plankton instance segmentation dataset, Bergum \emph{et.al} \cite{OCEANS}, there are only 131 images. This dataset contains 7 species of plankton and a total of 776 instances. As can be seen from Table~\ref{tab: comparison}, the ZMIS5K dataset is the first species-rich zooplankton microscopic image dataset. This indicates that the ZMIS5K dataset is more challenging for existing methods. Due to the significant variation in the area proportion of different zooplankton species at the same magnification level, we conducted further analysis on the ZMIS5K training set. The average size of the instances is 760,820 pixels, which is approximately 872×872 pixels, accounting for 15.2\% of the image size. Among these instances, those with an area smaller than 15.2\% of the image size account for 64.76\% of the total number of instances, while those with an area larger than 30\% of the image size account for 14.64\%.

\begin{figure}
\centering
\includegraphics[width=4.0in]{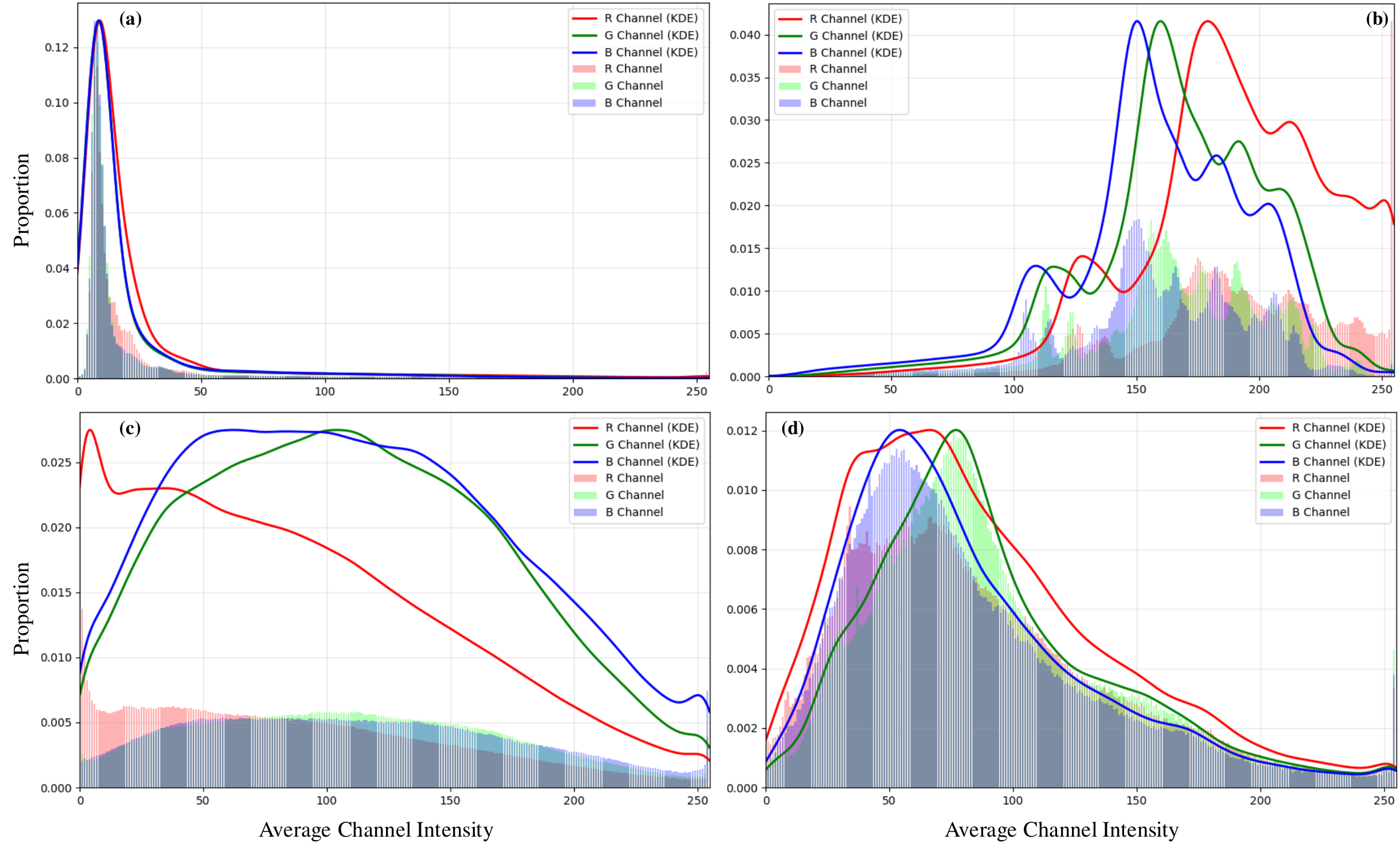}
\caption{Average channel intensity in the case of different datasets with proportion. (a) Average channel intensity of ZMIS5K under dark-field conditions. (b) Average channel intensity of ZMIS5K under flight-field conditions. (c) Average channel intensity of underwater images. (d) Average channel intensity of remote sensing images.}\label{fig: Intensity}
\end{figure}

\textbf{Channel Intensity of Zooplankton Images under Different Illumination Conditions.} Zooplankton in microscopic images exhibit significant differences in their characteristics under different illumination conditions. To quantify the differences between zooplankton microscopic images and images from other fields, we calculated the average channel intensity and probability density for each image separately on underwater images\cite{USIS-SAM}, remote sensing images \cite{7560644}, and the ZMIS5K dataset. As shown in the \cref{fig: Intensity}, the red channel in the ZMIS5K dataset has the highest density, while the R channel density in underwater images is the lowest. Similar to ZMIS5K, remote sensing images also have the highest density in the R channel, and this explains why RSPrompter achieves better segmentation performance on the ZMIS5K dataset than USIS-SAM. However, each channel of remote sensing images has only one peak, whereas zooplankton microscopic images have multiple peaks. Therefore, the RSPrompter model still poses significant challenges on the ZMIS5K dataset.

\section{Preliminaries}
\subsection{Preliminaries of SAM}
The SAM framework \cite{SAM} is an interactive prompt segmentation model that performs image segmentation tasks using prompt engineering, which comprises an image encoder, prompt encoder, and mask decoder.

\noindent\textbf{Image Encoder:} The image encoder is built on the pre-trained Vision Transformer (ViT) \cite{dosovitskiy2020vit}, which is composed of $L$ transformer layers, denoted $\mathcal{F}_{L}$. Given an input image $x \in \mathbb{R}^{H\times W\times 3}$, it is divided into $M=H\times W/ p^{2}$ fixed-size patches. After each patch is mapped to an embedding, it is concatenated with one class token for positional encoding, and then it is fed into the transformer layers:
\begin{equation}
\begin{aligned}
    x_{0} = [x_{cls}; \mathcal{E}_{M}(x)]
\end{aligned}
\end{equation}
\begin{equation}
\begin{aligned}
    \mathbf{F}_{img} = \mathcal{F}_{i}(x_{i-1}),\quad i= 1,..., L
\end{aligned}
\end{equation}
where $x_{cls}$ represents the class token for the global context. $\mathcal{E}_{M}(\cdot)$ represents the division of an image into $M$ patches and the projection to the embedding. $\mathcal{F}_{i} (\cdot)$ denotes $i$-th transformer layer.

\noindent\textbf{Prompt Encoder:} The prompt encoder maps prompts into this feature space, supporting both dense and sparse prompt inputs. Dense prompts refer to masks, which undergo convolution processing before being fused with image features. Sparse prompts can be arbitrary points, boxes, or text.
\begin{equation}
\begin{aligned}
    \mathbf{E}^{P}_{s} = \mathcal{P}_{s}(sparse)
\end{aligned}
\end{equation}
\begin{equation}
\begin{aligned}
    \mathbf{E}^{P}_{d} = \mathcal{P}_{d}(dense)
\end{aligned}
\end{equation}
where $\mathcal{P}_{s}(\cdot)$ and $\mathcal{P}_{d}(\cdot)$ represent the sparse prompt encoder and dense prompt encoder, respectively. $\mathbf{E}^{P}_{s}\in \mathbb{R}^{T\times C}$ represent sparse embedding, $T$ denotes the number of prompter token, and $C$ denotes the number of channels. $\mathbf{E}^{P}_{d}\in \mathbb{R}^{N\times C}$ represent dense embedding, $N$ denotes the spatial dimension.

\noindent\textbf{Mask Decoder:} The masked decoder, built on ViT with only 4M parameters, is a lightweight architecture. Integrates image features and prompt features using cross-attention to output the image mask. 
\begin{equation}
\begin{aligned}
   \{M_{k}, c_{k}\} = \mathcal{D}(\mathbf{F}_{img}, \mathbf{E}^{P}_{s}, \mathbf{E}^{P}_{d})
\end{aligned}
\end{equation}
where $M_k$ and $c_k$ represent mask logits and confidence, respectively. $\mathcal{D}(\cdot)$ denotes Mask Decoder.

\subsection{Preliminaries of Wavelet Transform}
As a multi-scale and multi-directional analysis tool, wavelet transform achieves precise analysis of local variations in zooplankton by matching wavelet functions at different scales and positions with the signal. During microscopic imaging, the frequency characteristics of the same zooplankton change under different lighting conditions. Wavelet transform can accurately analyze the variations in frequency characteristics within the image through local wavelet functions, thereby capturing subtle structural changes such as edges and textures of zooplankton.

\begin{figure}
\centering
\includegraphics[width=3.5in]{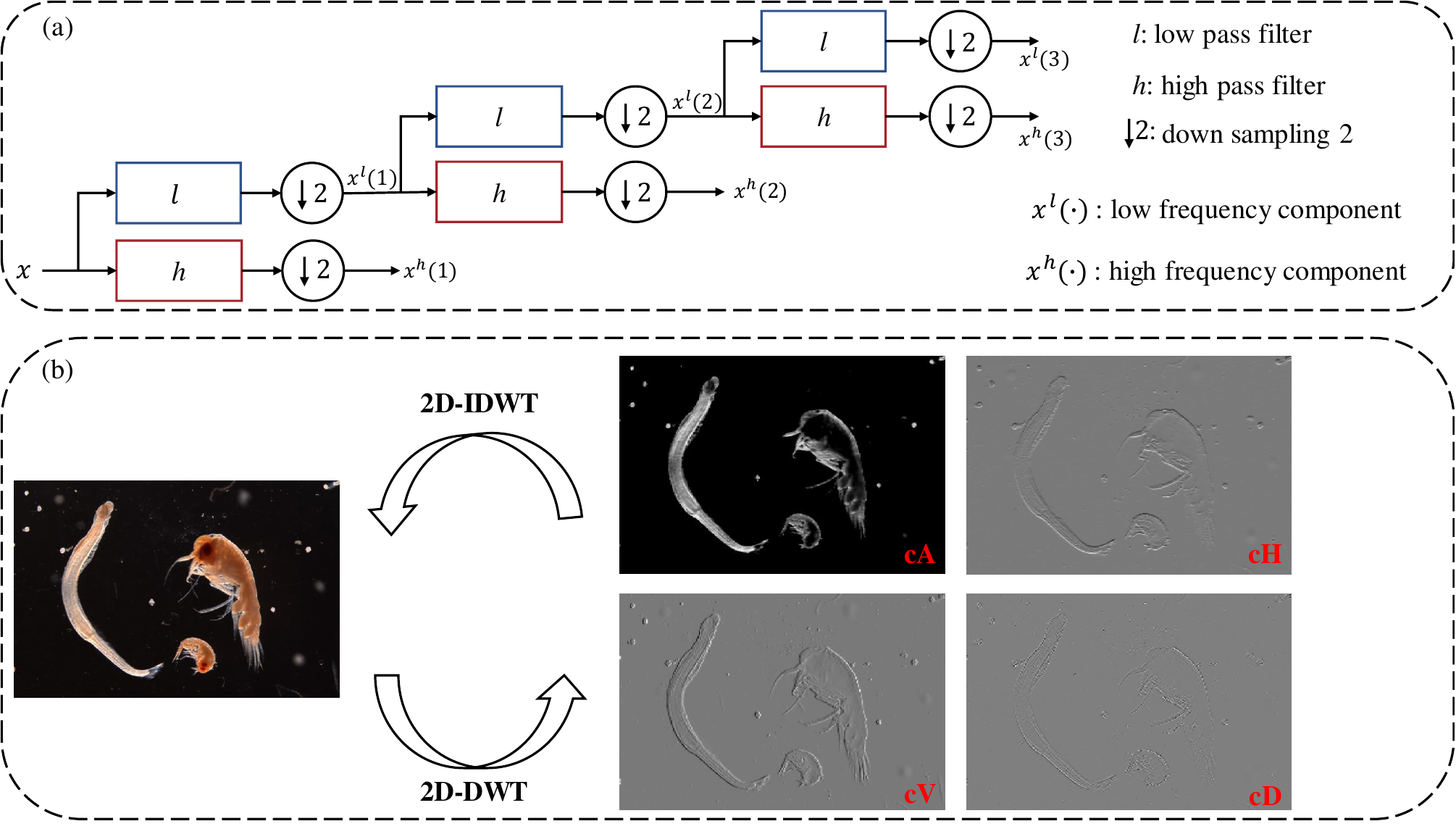}
\caption{Wavelet Transform Process. (a) The Multi-Scale Wavelet Transform Process. (b) The 2-Dimensional Discrete Wavelet Transform Process for Zooplankton. cA, cH, cV, and cD stand for Approximation Coefficients, Horizontal Detail Coefficients, Vertical Detail Coefficients, and Diagonal Detail Coefficients, respectively.}
\label{fig: WT}
\end{figure}

As shown in the \cref{fig: WT}(a), the discrete wavelet transform decomposes a signal into low‑frequency and high‑frequency components using low‑pass filters, high‑pass filters, and downsampling operations. In image processing, the image is regarded as a 2D discrete signal $I(x,y)$. The 2D discrete wavelet transform (2D-DWT) is performed by applying 1D discrete wavelet processing along the rows first and then the columns. The detailed formulas are given below:
\begin{equation}
\begin{aligned}
  L(x,y) = \sum_{k} I(k,y) l[2x - k]
\end{aligned}
\end{equation}
\begin{equation}
\begin{aligned}
  H(x,y) = \sum_{k} I(k,y) h[2x - k]
\end{aligned}
\end{equation}
where $L(x,y)$ and $H(x,y)$ denote the row-wise low-frequency and high-frequency signals of the image at pixel $(x,y)$, respectively, and $l$ and $h$ correspond to the low-pass and high-pass filters, respectively. $k$ represents the translation parameter, and $2x$ denotes 2× downsampling.

Next, we apply low-pass and high-pass filtering to $L$ and $H$ along the column direction of the image, respectively, ultimately yielding the following four frequency subbands of the image:
\begin{equation}
\begin{aligned}
  LL(x,y) = \sum_{k} L(x,k) l[2y - k]
\end{aligned}
\end{equation}
\begin{equation}
\begin{aligned}
  LH(x,y) = \sum_{k} L(x,k) h[2y - k]
\end{aligned}
\end{equation}
\begin{equation}
\begin{aligned}
  HL(x,y) = \sum_{k} H(x,k) l[2y - k]
\end{aligned}
\end{equation}
\begin{equation}
\begin{aligned}
  HH(x,y) = \sum_{k} H(x,k) h[2y - k]
\end{aligned}
\end{equation}
where $LL$ represents the low-frequency information of the image, corresponding to the overall contour structure of the zooplankton. $LH$, $HL$, and $HH$ denote the high-frequency information in the horizontal, vertical, and diagonal directions of the image, respectively, which correspond to the detailed information such as edges of the zooplankton.

As shown in the \cref{fig: WT}(b), after processing zooplankton microscopic images using the two-dimensional discrete wavelet transform, we can effectively extract the morphological structural features and edge detail features of zooplankton. This helps to compensate for the detail information lost during downsampling, thereby enhancing the model's ability to perceive fine-grained features of zooplankton.

\subsection{Loss function}
The loss function of ZMIS-SAM is similar to those in USIS-SAM and RSPrompter. It consists of localization loss $\mathcal{L}_{R}$, classification loss $\mathcal{L}_{C}$, bounding box loss $\mathcal{L}_{B}$, and segmentation loss $\mathcal{L}_{S}$, and the overall loss function is as follows.
\begin{equation}
\begin{aligned}
  \mathcal{L}_{R} = \beta_{1}\mathcal{L}_{C} + \theta_{1}\mathcal{L}_{B}
\end{aligned}
\end{equation}
\begin{equation}
\begin{aligned}
  \mathcal{L}_{T} = \alpha \mathcal{L}_{R} + \beta_{2}\mathcal{L}_{C} + \theta_{2}\mathcal{L}_{B} + \lambda\mathcal{L}_{S}
\end{aligned}
\end{equation}
where $\alpha$, $\beta_{1}$, $\beta_{2}$, $\theta_{1}$, $\theta_{2}$ and $\lambda$ denote the weight of each loss function in the total loss. Here, the weight of each loss function is set to 1.0. $\mathcal{L}_{R}$ denotes the region proposal loss in the RPN head, which consists of classification loss and bounding box loss. $\mathcal{L}_{C}$ denotes the cross-entropy loss, which is used to calculate the loss value between the ground-truth class and the predicted class. $\mathcal{L}_{B}$ denotes the SmoothL1Loss, which is used to calculate the loss value between the predicted bounding box and the ground-truth bounding box. $\mathcal{L}_{S}$ denotes the binary cross-entropy loss, which is used to calculate the loss value between the predicted segmentation mask and the ground-truth mask.

\subsection{Evaluation metrics}
We adopt mask Average Precision (AP), which is frequently used in a series of instance segmentation models such as Mask R-CNN, as the metric to evaluate model performance, including $mAP$, $AP_{50}$, and $AP_{75}$ under different Intersection over Union (IoU) thresholds. The AP can be measured by calculating the area under the precision-recall curve.
\begin{equation}
\begin{aligned}
  Precision = \frac{TP}{TP+FP}
\end{aligned}
\end{equation}
\begin{equation}
\begin{aligned}
  Recall = \frac{TP}{TP+FN}
\end{aligned}
\end{equation}
\begin{equation}
\begin{aligned}
  AP = \int_{0}^{1} P (r) , dr
\end{aligned}
\end{equation}
where $TP$, $FP$, and $FN$ denote the True Positive, False Positive, and False Negative, respectively. $P(r)$ denotes the precision at a recall of $r$.

\section{More ablation experiments of ZMIS-SAM}
\subsection{Effectiveness of the SA and IA in ZM-ViT}
We conduct an ablation study to evaluate the effectiveness of the Shape Adapter (SA) and Intensity Adapter (IA) in ZM‑ViT, with results summarized in \cref{tab: ablation-ia}. The SA is designed to steer the SAM encoder toward morphological structures of zooplankton, while the IA helps it adapt to channel‑wise intensity distributions in microscopic images. Experimental results confirm that both adapters consistently improve segmentation performance. As shown in \cref{tab: ablation-ia}, incorporating only SA or IA raises $mAP$ by 0.8\%  and 0.6\%, respectively. When both SA and IA are used together, $mAP$ increases from 70.4\% to 71.9\%, demonstrating that explicitly modeling zooplankton shape characteristics and intensity patterns effectively enhances SAM’s instance segmentation accuracy on zooplankton microscopy images.

\begin{minipage}[c]{0.4\textwidth}
    \centering
    \captionof{table}{\centering Ablation study of SA and IA.}
    \label{tab: ablation-ia}
    \setlength{\tabcolsep}{0.5 mm}
    \scalebox{0.9}{
    \begin{tabular}{lcccc}
    \hline
    SA & IA & $mAP$ & $AP_{50}$ & $AP_{75}$ \\\hline
      &  & 70.4 & 91.7 & 75.8 \\
    \textbf{$\checkmark$}  &  & 71.2 & 93.5 & 76.9 \\
       & \textbf{$\checkmark$} & 71.0 & 92.0 & 77.3 \\
    \textbf{$\checkmark$}  & \textbf{$\checkmark$} & \textbf{71.9} & \textbf{93.1} & \textbf{79.0}\\\hline
    \end{tabular}
    }
\end{minipage}
\hfill
\begin{minipage}[c]{0.6\textwidth}
    \captionof{table}{\centering Ablation study of optimal number of MConvBlocks in the NFAM.}
    \label{tab: ablation-n}
    \centering
    \setlength{\tabcolsep}{0.5 mm}
    \scalebox{0.9}{
    \begin{tabular}{lccc}
    \hline
    $n$ & $mAP$ & $AP_{50}$ & $AP_{75}$ \\\hline
    1   & 72.2 & 93.5     & 78.0 \\
    2   & 72.3 & 93.6     & 79.4 \\
    4   & \textbf{73.0} & 94.1 & \textbf{79.8}\\
    6   & 72.7 & \textbf{94.2} & 79.1\\\hline
    \end{tabular}
    } 
\end{minipage}

\subsection{Optimal Number of MConvBlocks in the NFAM}
We systematically evaluate how the number of MConvBlocks in the NFAM module affects the instance segmentation performance of ZMIS-SAM, with full results provided in \cref{tab: ablation-n}. The role of the MConvBlock is to extract multi-scale information from the aggregated features, thereby enhancing model robustness. We experiment with four values of $n$: 1, 2, 4, and 6. As shown in \cref{tab: ablation-n}, when $n$=1, the $mAP$ reaches 72.2\%. Performance improves gradually as $n$ increases, peaking at $n$=4. At $n$=6, both $mAP$ and $AP_{75}$ are slightly lower than those at $n$=4, while only $AP_{50}$ is marginally higher by 0.1\%, which falls within the range of normal variation. Based on these results, we select $n$=4 as the optimal number of MConvBlocks in the NFAM module.

\subsection{Effectiveness of the WM2FE}
WM2FE is a dynamic and learnable method based on the wavelet transform. We incorporate it into the USIS-SAM and train on the USIS10K, $mAP$ is further improved after incorporating WM2FE (\cref{tab: ablation-wm2fe}).
\begin{table}
\centering
\caption{\label{tab: ablation-wm2fe}The validated of WM2FE on the USIS10K dataset.}
\begin{tabular}{lcc|ccc|ccc}
\hline
Model & Epoch & Size & $mAP$ & $AP_{50}$ & $AP_{75}$ & $AP_{s}$ & $AP_{m}$ & $AP_{l}$\\\hline
USIS-SAM  & 24 & 512*512 & 34.6 & 51.7 & 39.1 & 11.5 & 27.7 & 38.6 \\
+WM2FE  & 24 & 512*512 & 35.4\color{red}{(+0.8)} & 52.6\color{red}{(+0.9)} & 40.7\color{red}{(+1.6)} & 8.7 & 27.6 & 40.2\color{red}{(+1.6)} \\\hline
\end{tabular}
\end{table}

\subsection{Parameter analysis of the model}
As shown in the \cref{tab: ablation-params}, we conduct a detailed analysis of the parameter quantities of each module in ZMIS-SAM. In the image encoder of ZMIS-SAM, from the eighth layer to the final layer, the original ViT blocks are replaced by ZM-ViT every two layers. Consequently, ZMIS-SAM contains 13 ZM-ViT blocks, 13 SA modules, 26 IA modules, and 13 NFAM modules in total. The number of trainable parameters for each SA, IA, and NFAM module are 1.65M, 1.23M, and 4.65M, respectively. Further analysis in \cref{tab: ablation-ia} shows that, by introducing two lightweight adapters (SA and IA) compared with the original SAM, ZMIS-SAM improves the instance segmentation metrics $mAP$, $AP_{50}$, and $AP_{70}$ by 1.5\%, 1.4\%, and 3.2\%, respectively. The WM2FE module has the smallest number of parameters and has a negligible impact on the overall trainable parameters. In summary, ZMIS-SAM enhances the instance segmentation accuracy of zooplankton microscopic images while maintaining a controllable increase in trainable parameters. Notably, under higher IoU thresholds, the model demonstrates more effective instance segmentation performance for zooplankton.
\begin{table}
\centering
\caption{\label{tab: ablation-params}The parameter count of each module, where $\dag$ represents the  trainable parameters of the module in ZMIS-SAM. SA refers to Shape Adapter, and IA refers to Intensity Adapter. NFAM refers to Neighboring Feature Aggregation Module, and WM2FE refers to Wavelet-based Multi-scale and Multi-directional Feature Enhancement.}
\begin{tabular}{lcccc}
\hline  
Module & SA$^{\dag}$ & IA$^{\dag}$ & NFAM$^{\dag}$ & WM2FE$^{\dag}$ \\\hline
Params  & 21.4M & 32.0M & 60.0M & 0.5M\\\hline
\end{tabular}
\end{table}

\subsection{Analysis of Model Inference Speed}
Zooplankton analysis is typically offline (not real-time). ZMIS-SAM can adequately process samples in laboratory environments at 3.9 FPS and does not occupy excessive GPU memory during inference (\cref{tab: ablation-inference})
\begin{table}
\centering
\caption{\label{tab: ablation-inference}Model inference efficiency on the same hardware.}
\begin{tabular}{lcccccc}
\hline  
Method & FPS & Latency & CUDA Mem & $mAP$ & $AP_{50}$ & $AP_{75}$\\\hline
RSPrompter  & 8.8 & 113.6 & 2596M & 71.8 & 92.2 & 76.9\\
ZMIS-SAM  & 3.9 & 256.4 & 2939M & 73.6\color{red}{(+1.8)} & 94.6\color{red}{(+2.4)} & 80.7\color{red}{(+3.8)}\\\hline
\end{tabular}
\end{table}

\subsection{Effectiveness analysis of each model}
\begin{figure}
\centering
\includegraphics[width=4.0in]{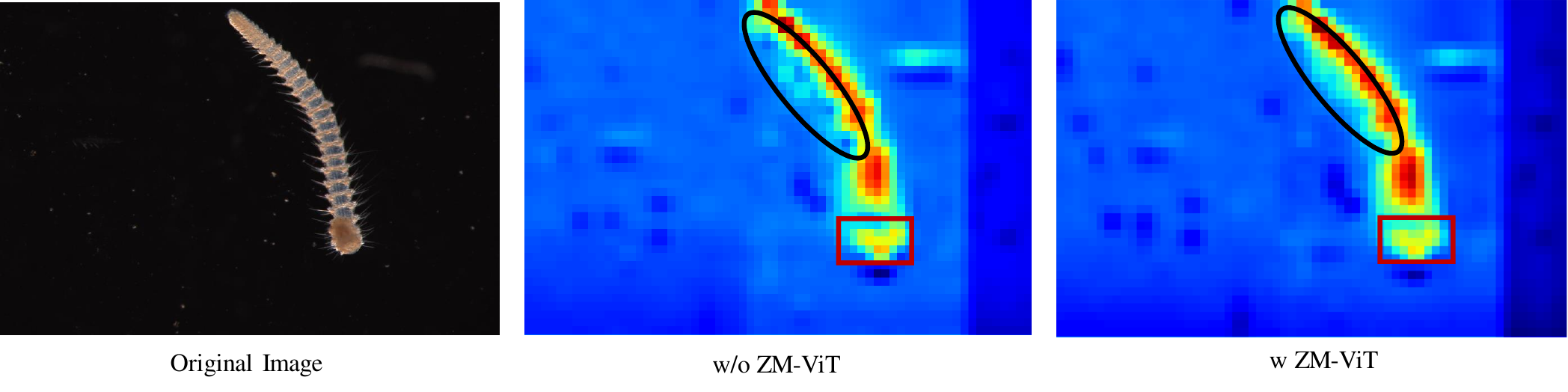}
\caption{Visualize the feature map of ZM-ViT. The ZM-ViT enables SAM to learn knowledge of zooplankton.}
\label{fig: zmvit_map}
\end{figure}
In this section, we further analyze the effectiveness of the three modules (ZM-ViT, NFAM, and WM2FE) through feature map visualization. As shown in \cref{fig: zmvit_map}, we can clearly observe that the features extracted by the model exhibit discernible refinement in both spatial structure and channel-wise response patterns after processing through ZM-ViT. This improvement stems from the dedicated roles of its two components: 1) Shape Adapter guides SAM to focus on morphological characteristics of zooplankton, thereby enhancing the model's capacity to represent shape variations. 2) Intensity Adapter enables SAM to adaptively adjust its response to channel-wise intensity distributions, more accurately capturing illumination variations under microscopic imaging conditions. Working synergistically, these adapters not only strengthen the model's feature representation in key regions but also effectively suppress responses to background impurities.

\begin{figure}
\centering
\includegraphics[width=4.0 in]{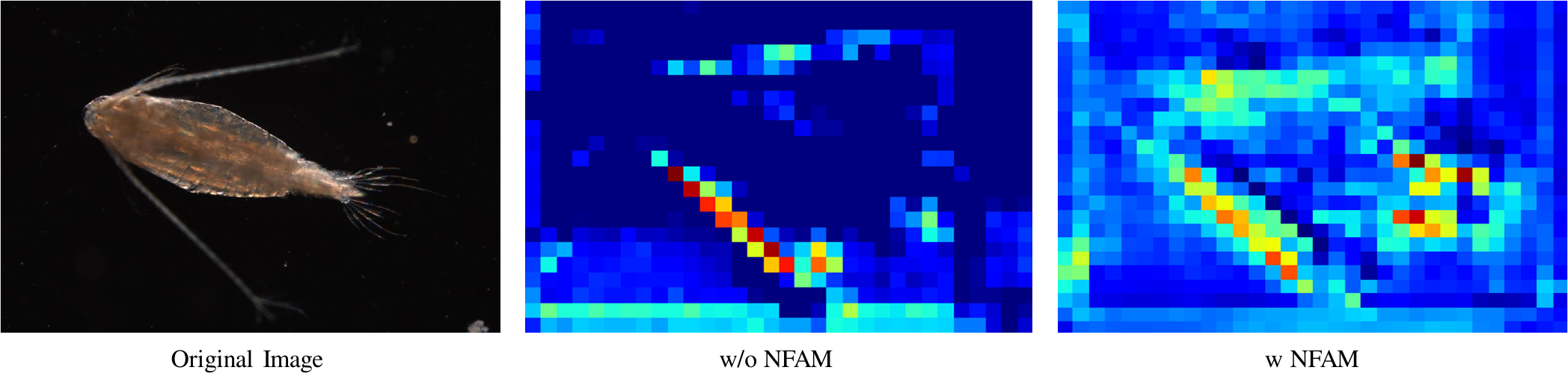}
\caption{Visualize the feature map of NFAM. The NFAM enhances the model's ability to continuously segment slender appendages.}
\label{fig: nfam_map}
\end{figure}
\cref{fig: nfam_map} clearly illustrates through feature visualization how the NFAM effectively integrates general-purpose features extracted by the ViT layers with domain-specific features learned through ZM-ViT, establishing semantic associations between the main body structures of zooplankton and their slender appendages in the feature space. Furthermore, in the model without NFAM, the feature responses in the slender appendage regions appear significantly fragmented and discontinuous, preventing the model from perceiving the complete anatomical structure. In contrast, the NFAM successfully captures the progressive feature variations along these delicate appendages, thereby substantially enhancing the model's capability for continuous segmentation of such structures.

\begin{figure}
\centering
\includegraphics[width=4.0 in]{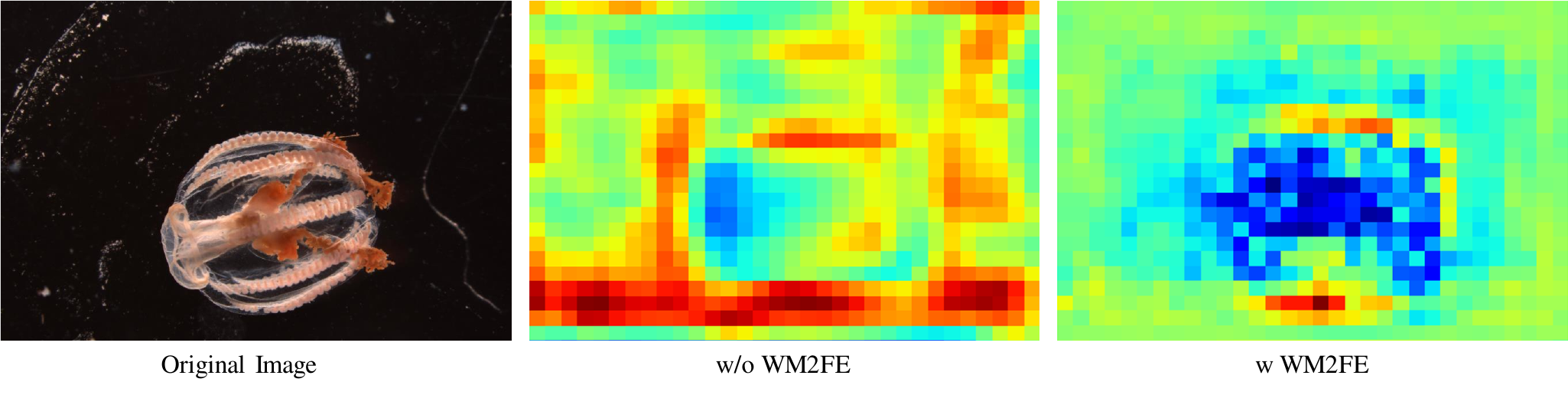}
\caption{Visualize the feature map of WM2FE. The WM2FE focuses more on species boundaries, thus enhancing the model's segmentation ability for boundaries. }
\label{fig: wm2fe_map}
\end{figure}

As shown in \cref{fig: wm2fe_map}, comparative feature map visualizations clearly demonstrate that incorporating the WM2FE module significantly enhances the model's feature representation of zooplankton boundaries. Without WM2FE, the boundary features of \emph{Pleurobrachia globosa} appear blurred and poorly localized, particularly along its transparent edge structures. In contrast, the WM2FE module employs the Haar wavelet transform to extract high-frequency details along vertical, horizontal, and diagonal orientations, effectively compensating for the loss of fine details during upsampling. This integration of multi-directional high-frequency information substantially strengthens the model's capacity to represent and segment complex boundary structures in zooplankton specimens.

\subsection{Clear comparative visualization results}
\begin{figure}
\centering
\includegraphics[width=5.0in]{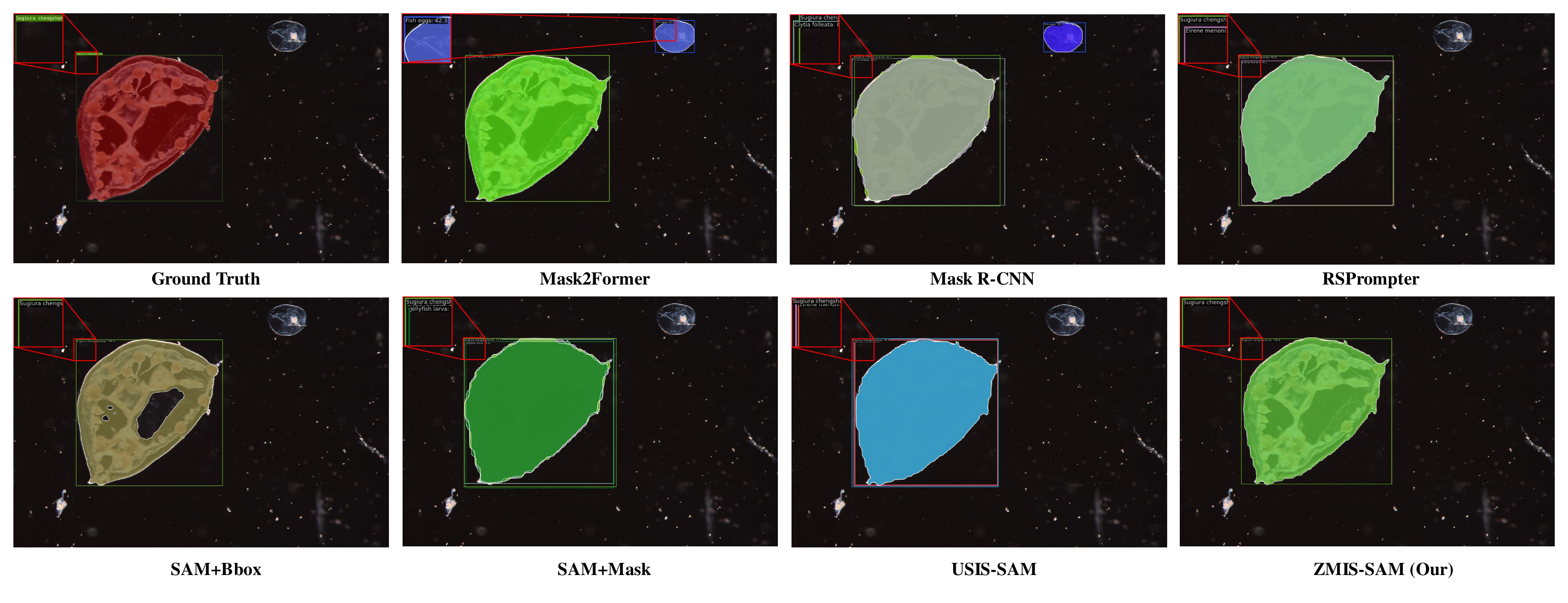}
\caption{Visualization results of inaccurate classification cases.}
\label{fig: 5-1}
\end{figure}

\begin{figure}
\centering
\includegraphics[width=5.0in]{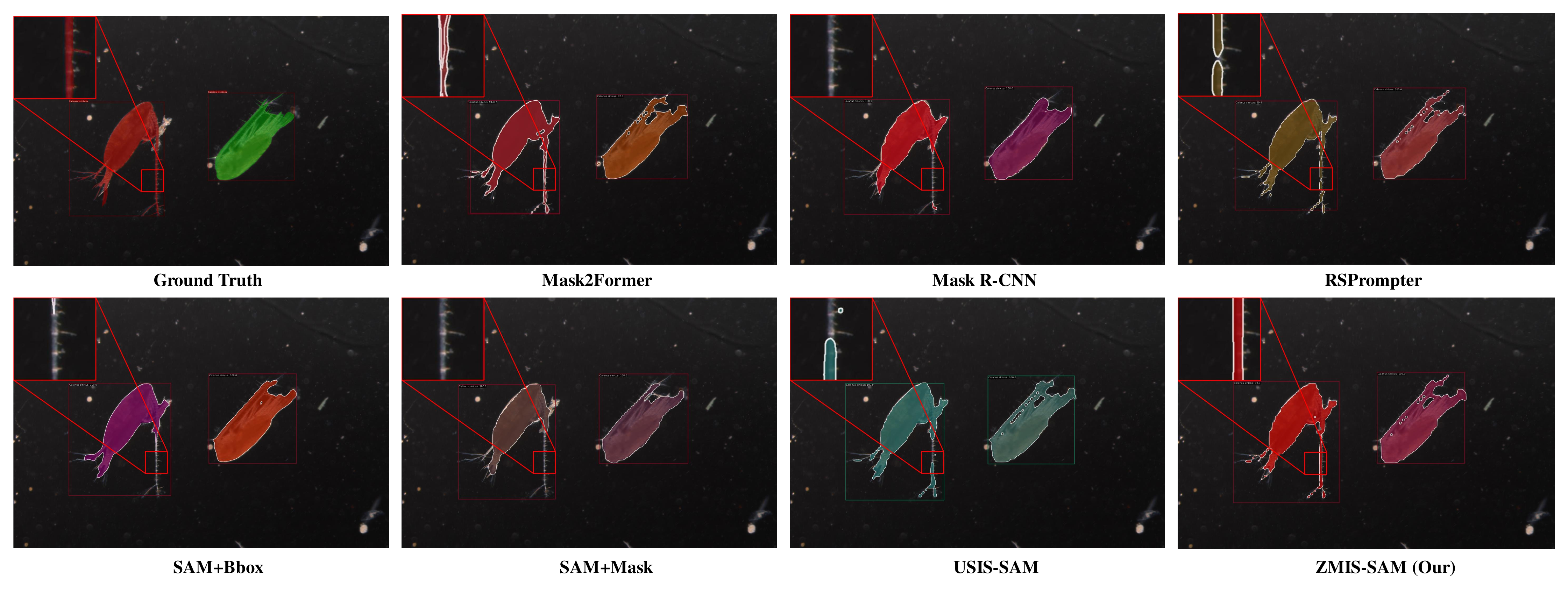}
\caption{Visualization results of discontinuous segmentation of slender appendages cases.}
\label{fig: 5-2}
\end{figure}

\begin{figure}
\centering
\includegraphics[width=5.0in]{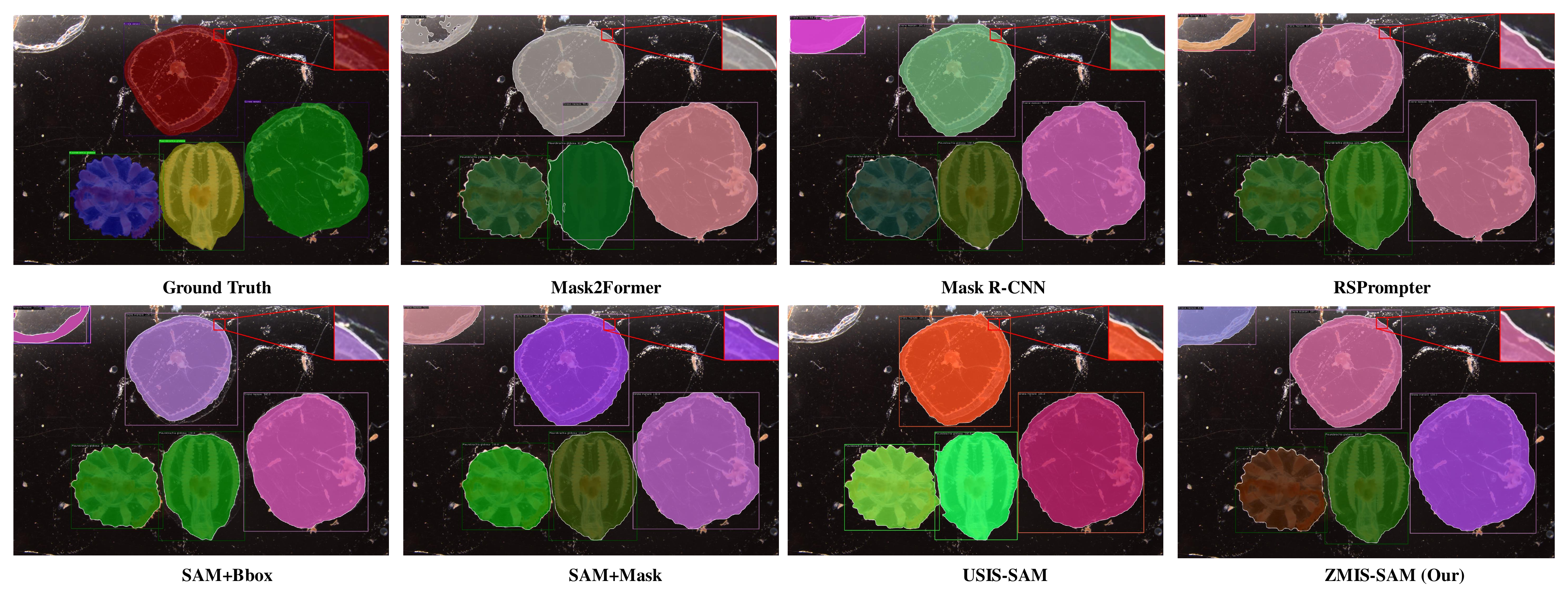}
\caption{Visualization results of incomplete boundary segmentation cases.}
\label{fig: 5-3}
\end{figure}

In this subsection, we present clearer visualizations compared to those shown in the main text (\cref{fig: 5-1}, \cref{fig: 5-2}, and \cref{fig: 5-3}).

\section{Generalization experiments}
This section evaluates the cross-domain instance segmentation performance of the proposed model. Due to the absence of publicly available instance segmentation datasets for zooplankton or even plankton microscopy images in general, we retrain ZMIS-SAM on commonly adopted public benchmarks originally used for SAM-based instance segmentation to validate its generalization ability across specialized domains.

\subsection{Datasets}
UIIS \cite{WaterMask} is an underwater instance segmentation dataset with a total of 4,628 images, covering 7 underwater instance categories: Fish, Reefs, Aquatic plants, Wrecks/ruins, Human divers, Robots, and Sea-floor. It was split into 3,937 training images and 691 validation images. NWPU \cite{7560644} is a remote sensing instance segmentation dataset consisting of 650 images in total, including 10 remote sensing instance categories: cars, airplanes, ships, etc. It was divided into 454 training images and 196 validation images.

\subsection{Implementation Details}
For fair comparison, both USIS-SAM and ZMIS-SAM use an input size of 512×512, an initial learning rate of 1e-4, and a batch size of 12. Following the experimental setup in USIS-SAM, the models are trained for 30 epochs on UIIS and 100 epochs on NWPU. To further assess sensitivity to objects at different scales, we include three additional metrics: $AP_{s}$, $AP_{m}$, and $AP_{l}$, evaluating performance on small, medium, and large targets, respectively.

\subsection{Experiments Results}
As shown in \cref{tab: ablation-Gener}, ZMIS-SAM achieves $mAP$, $AP_{50}$, and $AP_{75}$ scores of 34.1\%, 60.9\%, and 30.0\% on the NWPU dataset, significantly outperforming USIS-SAM and demonstrating stronger cross-domain generalization. On the UIIS dataset, ZMIS-SAM also surpasses USIS-SAM in mAP, $AP_{s}$, $AP_{m}$, and $AP_{l}$, with a notable 1.6\% improvement in $AP_{l}$, indicating its superior performance in segmenting large objects. However, it should be noted that USIS-SAM, which is specifically designed for underwater imagery, achieves higher $AP_{50}$ and $AP_{75}$ in this domain, reflecting its specialized strength in certain localization tasks.

\begin{table*}
\centering
\caption{\label{tab: ablation-Gener} Quantitative comparisons with USIS-SAM on the UIIS and the NWPU.}
\begin{tabular}{lcccc|ccc}
\hline
Datasets & Method & $mAP$ & $AP_{50}$ & $AP_{75}$ &$AP_{s}$ & $AP_{m}$ & $AP_{l}$\\\hline
NWPU \cite{7560644} & USIS-SAM \emph{(ICML'24)} \cite{USIS-SAM} & 16.1 & 30.0 & 16.1 & 3.5 & 9.9 & 35.1\\
 & ZMIS-SAM & \textbf{34.1} & \textbf{60.9} & \textbf{30.0} & \textbf{8.6} & \textbf{28.2} & \textbf{57.7}\\\hline
UIIS \cite{WaterMask} & USIS-SAM \emph{(ICML'24)} \cite{USIS-SAM} & 22.0 & 36.9 & 23.5 & 4.8 & 17.2 & 33.2\\
 & ZMIS-SAM & \textbf{22.5} & 36.7 & 23.4 & \textbf{5.3} & \textbf{17.2} & \textbf{34.8}\\\hline
\end{tabular}
\end{table*}

\end{document}